\documentclass{article}

\PassOptionsToPackage{numbers}{natbib}

\usepackage[final]{neurips_2022}

\usepackage[utf8]{inputenc} 
\usepackage[T1]{fontenc}    
\usepackage{hyperref}       
\usepackage{url}            
\usepackage{booktabs}       
\usepackage{amsfonts}       
\usepackage{nicefrac}       
\usepackage{microtype}      
\usepackage{xcolor}         
\usepackage{enumitem}
\usepackage{amsmath}
\usepackage{multirow}
\usepackage{float}
\usepackage{bbm}
\usepackage{graphicx}
\usepackage{algorithm}
\usepackage{algcompatible}
\usepackage{algpseudocode}
\algtext*{EndWhile}
\algtext*{EndFor}
\algtext*{EndIf}
\usepackage{caption}
\algnewcommand\algorithmicforeach{\textbf{for each}}
\algdef{S}[FOR]{ForEach}[1]{\algorithmicforeach\ #1\ \algorithmicdo}
\DeclareMathOperator*{\argmax}{arg\,max}
\DeclareMathOperator*{\argmin}{arg\,min}

\usepackage[utf8]{inputenc} 
\usepackage[T1]{fontenc}    
\usepackage{hyperref}       
\usepackage{url}            
\usepackage{booktabs}       
\usepackage{amsfonts}       
\usepackage{nicefrac}       
\usepackage{microtype}      
\usepackage{xcolor}         
\usepackage{enumitem}
\usepackage{amsmath}
\usepackage{multirow}
\usepackage{float}
\usepackage{bbm}
\usepackage{graphicx}
\usepackage{algorithm}
\usepackage{algcompatible}
\usepackage{algpseudocode}
\algtext*{EndWhile}
\algtext*{EndFor}
\algtext*{EndIf}
\usepackage[font=footnotesize]{caption}
\usepackage{soul}

\algdef{S}[FOR]{ForEach}[1]{\algorithmicforeach\ #1\ \algorithmicdo}

\title{PALMER: Perception-Action Loop with Memory \\ for Long-Horizon Planning}

\author{%
  Onur Beker \\
  \And
  Mohammad Mohammadi\\
  \And
  Amir Zamir \\
  \AND
  \normalfont{Swiss Federal Institute of Technology (EPFL)} \\
}

\begin{document}
\maketitle

\begin{abstract}
To achieve autonomy in a priori unknown real-world scenarios, agents should be able to: i) act from high-dimensional sensory observations (e.g., images), ii) learn from past experience to adapt and improve, and iii) be capable of long horizon planning. 
Classical planning algorithms (e.g. PRM, RRT) are proficient at handling long-horizon planning. Deep learning based methods in turn can provide the necessary representations to address the others, by modeling statistical contingencies between observations. 
In this direction, we introduce a general-purpose planning algorithm called PALMER that combines classical sampling-based planning algorithms with learning-based perceptual representations. For training these perceptual representations, we combine Q-learning with contrastive representation learning to create a latent space where the distance between the embeddings of two states captures how easily an optimal policy can traverse between them. For planning with these perceptual representations, we re-purpose classical sampling-based planning algorithms to retrieve previously observed trajectory segments from a replay buffer and restitch them into approximately optimal paths that connect any given pair of start and goal states. This creates a tight feedback loop between representation learning, memory, reinforcement learning, and sampling-based planning. The end result is an experiential framework for long-horizon planning that is significantly more robust and sample efficient compared to existing methods. 
\end{abstract}

\section{Introduction}
\label{introduction}

\begin{figure}[t]
  \centering
  \includegraphics[width=\textwidth]{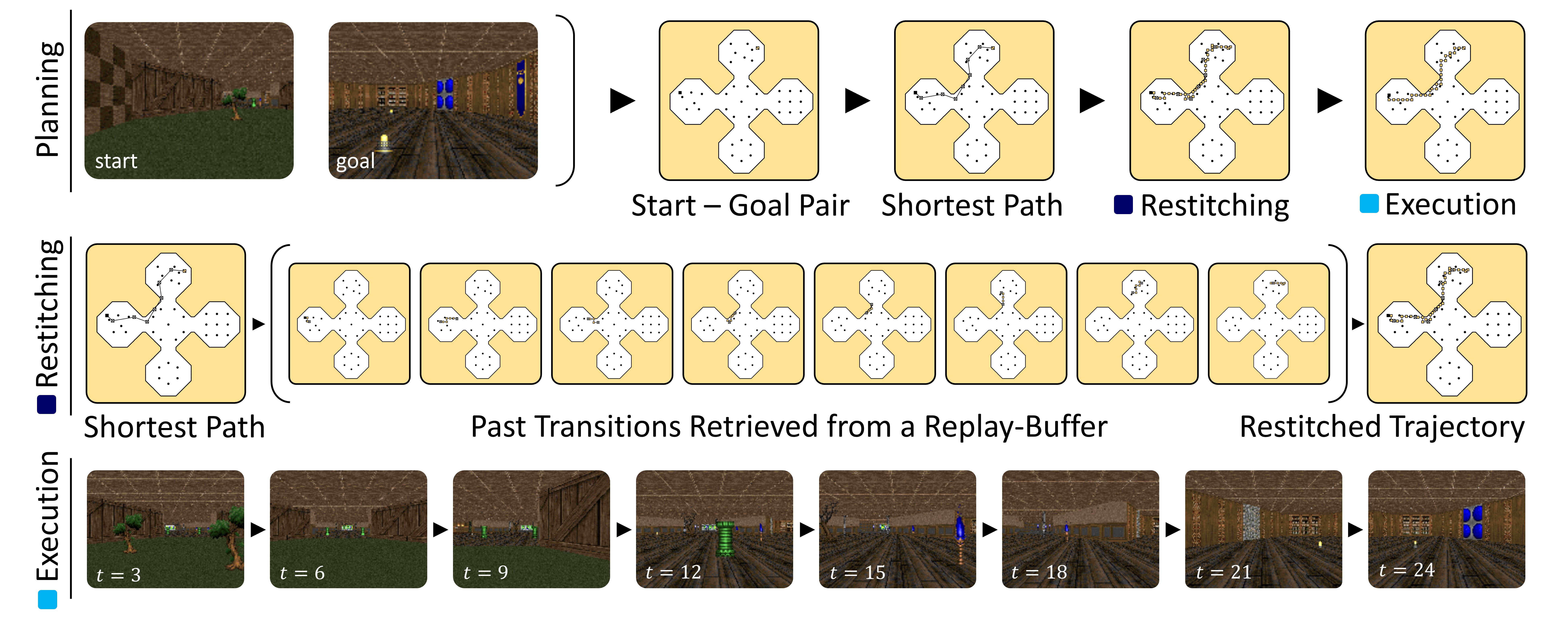}
  \includegraphics[width=\textwidth]{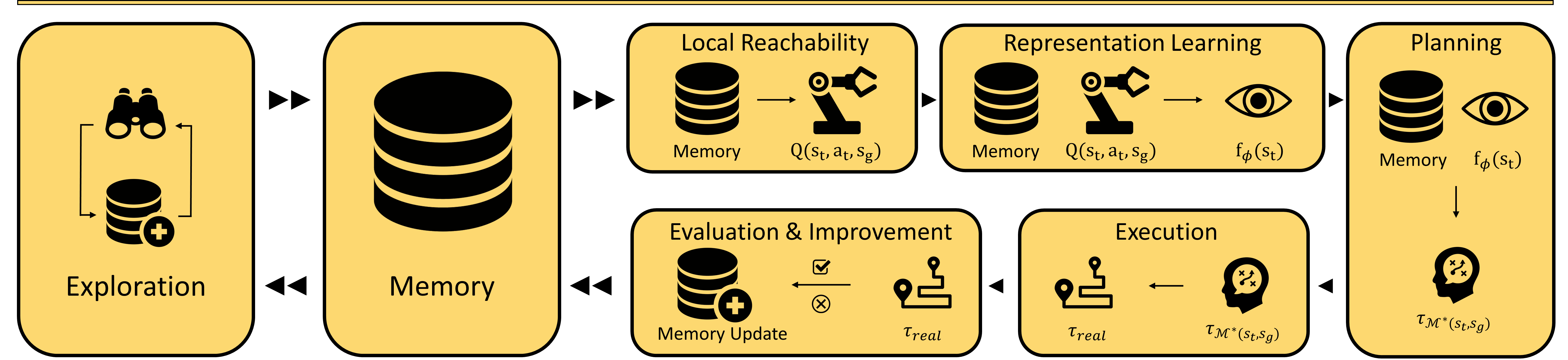}
  \vspace*{-\baselineskip}
  \caption{\textbf{Top}: Given a start-goal image pair, PALMER plans a path between them by concatenating the endpoints of past trajectory segments retrieved from a provided replay buffer. This is enabled by a state embedding function $f_{\phi}$ that can identify close-by states, and results in robust long-horizon planning. \textbf{Bottom}: To achieve this : i) it uses offline Q-learning to obtain \textit{local reachability} estimates between states, ii) uses these Q-values for \textit{representation learning} to train $f_{\phi}$, iii) uses $f_{\phi}$ to \textit{plan} over the replay buffer, iv) \textit{executes} these plans, v) \textit{evaluates} the resulting trajectories and inserts them back into the replay buffer to \textit{improve} its contents.}
  \vspace*{-\baselineskip}
  \vspace*{-1.5mm}
  \label{fig:method}
\end{figure}

Animals and humans operate on high-dimensional stimuli (e.g., vision) to achieve diverse and ever-changing goals necessary for their survival \cite{palmer1999vision, gibson2014ecological, o2001sensorimotor, co2020ecological, silver2021reward}. 
Learning through trial-and-error plays a fundamental role in this \cite{thorndike1898animal, thorndike1927law, campbell1956perception, herrnstein1970law, whitehead1991learning, silver2021reward}. Even in simplest environments, a brute-force approach to trial-and-error by trying every possible action for achieving every possible goal is intractable. The complexity of this search motivates memory-based mechanisms for compositional thinking. Examples of such mechanisms include : i) remembering relevant segments of past experience, ii) recomposing them in new counterfactual ways to form plans, and iii) executing such plans as part of a targeted search strategy. Such mechanisms for recycling past successful behavior can significantly accelerate trial-and-error compared to uniformly sampling all possible actions. This is because the same behavior (i.e., sequence of actions) can remain valid for different goals and in different contexts, due to the inherent compositional structure of real-world goals as well as the commonality of the physical laws that govern real-world environments.  

What principles can allow for memory mechanisms to remember and recompose bits of experience?
The concept of dynamic programming (DP) is directly related to this discussion, as it greatly reduces the computational cost of trial-and-error by employing the principle of optimality \cite{bertsekas2012dynamic}. This principle can be colloquially stated as treating new and complex problems as a recomposition of old and simpler sub-problems that were already solved before. Recent work \cite{eysenbach2019search, emmons2020sparse, savinov2018semi} employs this perspective to build hierarchical reinforcement learning (RL) algorithms for goal-reaching tasks. Such methods set edges between states using a distance regression model to build a planning graph, perform shortest path computations over it using DP-based graph search, and follow the resulting shortest paths with a learning-based local policy. Our paper builds upon this line of work.

\underline{\textbf{\textit{Contribution:}}} We describe a long-horizon planning method that directly operates on high dimensional sensory input observable by an agent on its own (e.g., images from an onboard camera). Our method combines classical sampling-based planning algorithms with learning-based perceptual representations, to retrieve and recompose previously observed sequences of state transitions in a replay buffer. This is enabled by a two-step process. \textit{First}, we learn a latent space where the distance between two states captures how many timesteps it takes for an optimal policy to go from one to the other. To achieve this, we use goal-conditioned Q-values learned through offline hindsight relabelling \cite{andrychowicz2017hindsight} for contrastive representation learning. 
\textit{Second}, we threshold this learned latent distance metric to define a neighborhood criterion between states. We then define sampling-based planning algorithms that search over the replay buffer \cite{eysenbach2019search} to retrieve and stitch together trajectory segments (i.e., past sequences of observed transitions) whose endpoints are neighboring states. This trajectory stitching approach allows for creating planning graphs to connect any pair of start and goal states that were observed before (as depicted in Fig.\ref{fig:method}). Our approach operates on offline unlabeled data, and can therefore be combined with any exploration method to populate the replay buffer.
Our experiments implement an image-based navigation policy in simulation, using an offline replay buffer populated with uniform random-walk exploration data.

\section{Perception-Action Loop with Memory Retrieval\protect\footnote{Most sub-sections have a corresponding section in the supplementary for further elaboration.}}
\label{PER}

\underline{\textit{Nomenclature:}} An environment is represented as a tuple $\langle \mathcal{S}, \mathcal{A}, p_{env} \rangle$, where $\mathcal{S}$ and $\mathcal{A}$ are the state and action spaces, and $p_{env}(s'|s, a)$ is the Markovian transition dynamics. A trajectory $\tau \in \mathcal{T}$ is any sequence of states and actions. 
$\tau_{0}$ , $\tau_{-1}$ , $\tau_{i}$ denote the first, last, and $i$'th states in $\tau$ respectively. The length of a trajectory in terms of timesteps is denoted as $len(\tau)$, and concatenation of two trajectories is denoted as $\tau_{cat} = \tau_1 \circ \tau_2$.
We assume an additive reward function $\mathcal{R}: \mathcal{T} \rightarrow \mathbb{R}$ where $\mathcal{R}(\tau) = \sum_{(s, a) \in \tau}r(s,a)$.
We call a finite set of trajectories $\mathcal{M} = \{\tau_i\}$ a replay buffer.

\subsection{Perceptual Representations that Capture Local Reachability}
\label{sec:representations}
A key component of our framework is a perceptual encoder $f_{\phi}(s): \mathcal{S} \rightarrow \mathbb{R}^d$ that maps states into a representation space where L2 distance $d_{\phi}(s_t, s_g) := \|f_{\phi}(s_t) - f_{\phi}(s_g)\|$ captures local reachability (i.e., how many timesteps it takes for the optimal policy to go from one state to another). To discuss this more rigorously, we follow the work of \cite{kaelbling1993learning, eysenbach2019search} and define a goal-conditioned reward function $r(s_t, a, s_{t+1}, s_g) = - \mathbbm{1}_{s_{t+1} \neq s_g}$ that returns $-1$ for all steps before reaching a goal. This means goal-conditioned Q-values \cite{kaelbling1993learning, schaul2015universal} for the optimal policy correspond to negative shortest-path distances (i.e., $max_a Q(s_t, a, s_g) = V(s_{i}, s_{j}) = - len(\tau_{sp})$). We can then define a symmetric distance metric between states as $d_Q(s_{c}, s_{g}) := max ( \ - V(s_{c}, s_{g}), - V(s_{g}, s_{c}) \ )$. This in turn corresponds to the two-way consistency criterion proposed in \cite{emmons2020sparse}. What we want from $f_{\phi}(s)$ is for $d_{\phi}(s_{c}, s_{g})$ and $d_Q(s_{c}, s_{g})$ to roughly correlate.

\subsection{Representation Learning via Reinforcement Learning}
\label{sec:architecture}

\begin{figure}[t]
  \centering
  \includegraphics[width=\textwidth]{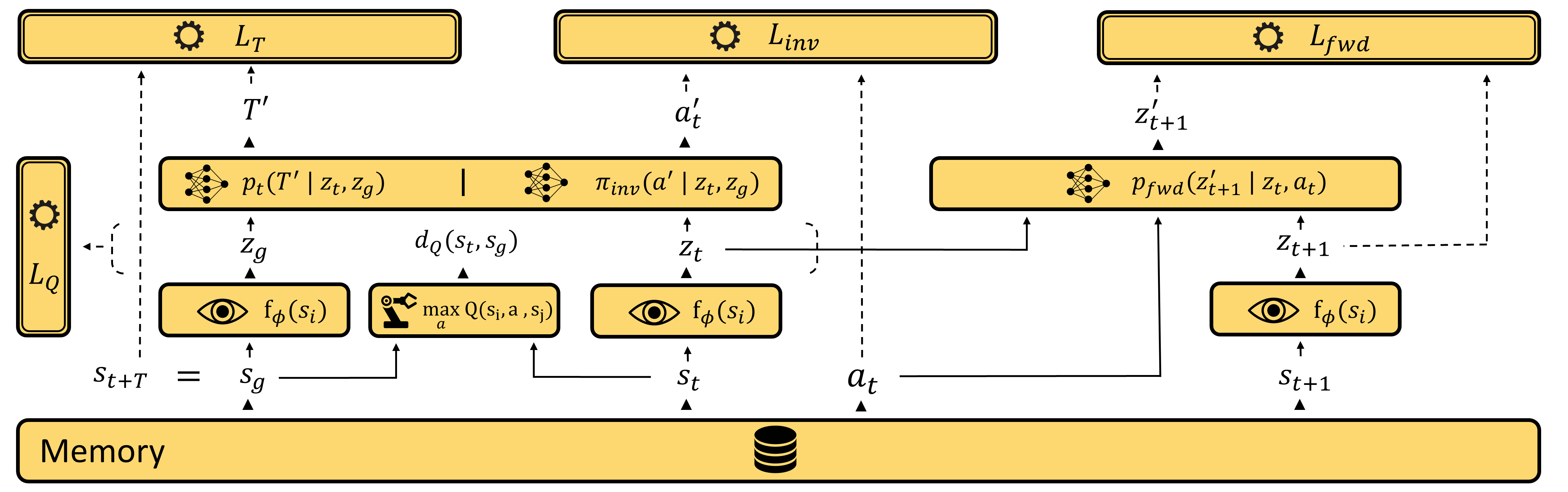}
  \vspace*{-\baselineskip}
  \caption{An overview of the functions, inputs, and losses used in our method (see Sec.\ref{sec:architecture} for details). We aim to train a perceptual encoder $f_{\phi}$ with two properties: i) representations of two states should be close if they were observed to be easily reachable from each other within a low number of timesteps, ii) the representation of a state should capture a minimal sufficient statistic to inform an agent about the actions needed to reach nearby states.}
  \label{fig:architecture}
  \vspace*{-\baselineskip}
\end{figure}

Any perceptual encoder $f_{\phi}$ whose latent representations satisfy the local reachability property defined in Sec.\ref{sec:representations} can be used to implement the nearest neighbor retrieval and trajectory stitching mechanisms for the upcoming sections \ref{sec:PER} and \ref{sec:retro_plan}. This section discusses \textit{one possible way} to obtain such a perceptual encoder, by using goal-conditioned Q-values for contrastive representation learning.

We propose a model (depicted in Fig.\ref{fig:architecture}) that includes the following standard components from the literature: \textbf{i)} $z = f_{\phi}(s)$, projecting a state into a latent representation; \textbf{ii)} $p_{fwd}(z'_{t+1} \ | \ z_t, a_t)$, modelling the transition distribution induced by $p_{env}(s'|s, a)$ over the latent space $z = f_{\phi}(s)$, as discussed in \cite{agrawal2016learning, pathak2017curiosity}; \textbf{iii)} $\pi_{inv}(a'_t \ | \ z_t, z_g)$, defining a distribution of actions to reach a goal state, as discussed in \cite{agrawal2016learning, pathak2017curiosity, savinov2018semi}; \textbf{iv)} $p_{t}(T' \ | \ z_t, z_g)$, modelling the distribution of timesteps necessary to reach a goal state, as discussed in \cite{hartikainen2019dynamical}; \textbf{v)} $Q(s_t, a_t, s_g)$, a Q-value function that provides local reachability estimates between pairs of states, as discussed in \cite{eysenbach2019search, kaelbling1993learning}. 

Following \cite{eysenbach2019search, kaelbling1993learning}, we train $Q(s_t, a_t, s_g)$ over an offline replay buffer $\mathcal{M}$, using hindsight relabelling \cite{andrychowicz2017hindsight, kaelbling1993learning} with a reward function $r(s_t, a, s_{t+1}, s_g) = - \mathbbm{1}_{s_{t+1} \neq s_g}$.
After training $Q(s_t, a_t, s_g)$ in isolation, we freeze its parameters and use it to define a contrastive loss function \cite{chopra2005learning} $L_Q$ as explained below. We then train the remaining components using the same replay buffer $\mathcal{M}$. We randomly sample a transition $(s_t, a_t, s_{t+1})$ and a time difference $T$, and set the goal state as $s_g := s_{t+T}$, as in hindsight relabelling. We then minimize the following losses: 

\begin{itemize}[leftmargin=*]
\item $L_Q(s_t, s_g) = l_{hinge}(d_{\phi}(s_t, s_g) - d_{p}) \ \mathbbm{1}_{d_Q(s_t, s_g) \leq c_Q} + l_{hinge}(d_{p} - d_{\phi}(s_t, s_g) ) \ \mathbbm{1}_{d_Q(s_t, s_g) \ge c_Q}$, where $l_{hinge}$ is the hinge loss \cite{shalev2014understanding}. This contrastive loss dictates that perceptual representations should be close together (i.e., $d_{\phi}(s_t, s_g) \leq d_{p}$ holds) if and only if two states are close to each other in terms of reachability (i.e., $d_Q(s_t, s_g) \leq c_Q$ holds). $d_p$ and $c_Q$ are hyperparameters. 
\item $L_T(T', T)$, $L_{inv}(a_t', a_t)$, and $L_{fwd}(z_{t+1}', z_{t+1})$ are MSE and cross-entropy losses \cite{pathak2017curiosity, hartikainen2019dynamical}. $L_T$ and $L_{inv}$ dictate that perceptual representations should capture enough information to know when and how an agent can reach from one state to another, while $L_{fwd}$ dictates that they should capture only a minimal-sufficient statistic for doing so (\cite{pathak2017curiosity} presents a more elaborate discussion).
\end{itemize}

\subsection{Perceptual Experience Retrieval (PER)} 
\label{sec:PER}
Given a perceptual encoder $f_\phi$ that captures local reachability, we go over all states ${s_i} \in \mathcal{M}$ in the replay buffer and compute their projections ${z_i = f_\phi(s_i)}$, which are stored alongside the states themselves. We then employ $z_i$ to implement two retrieval mechanisms from the replay buffer: i) retrieving neighboring states, and ii) retrieving neighboring trajectories. 

\underline{\textit{i) Retrieving Neighboring States:}} Given a query state $s_c$ and radius $d_p$ (i.e., the same one used in the contrastive loss $L_Q$ in Sec.\ref{sec:architecture}), retrieving neighboring states amounts to computing the set $\mathcal{N}_{d_p}(s_c) = \{s_n \ | \ d_\phi(s_c, s_n) \leq d_p\}$, which can be achieved by a straightforward L2 distance computation and thresholding. 
The number of neighbors $|\mathcal{N}_{d_p}(s_c)|$ of a query state $s_c$ is an approximate measure of how many times the agent has visited around $s_c$, which also makes it a good visitation-count that is applicable to both discrete and continuous state spaces.


\underline{\textit{ii) Retrieving Neighboring Trajectories:}} Given a starting state $s_c$ and a goal state $s_g$, we can search the replay buffer for the highest reward trajectory segment $\tau$ that starts from a state $\tau_{0}$ in $\mathcal{N}_{d_p}(s_c)$ and ends in a state $\tau_{-1}$ in $\mathcal{N}_{d_p}(s_g)$. This corresponds to the following optimization problem:
\begin{equation} \label{eq:PER}
    \tau_{\mathcal{M}(s_c, s_g)} := \argmax_{\tau \in \mathcal{M}}{\mathcal{R}(\tau)} \quad \textrm{s.t.} \quad \tau_{0} \in \mathcal{N}_{d_p}(s_c) \ , \ \tau_{-1} \in \mathcal{N}_{d_p}(s_g)
\end{equation}

To find $\tau_{\mathcal{M}(s_c, s_g)}$, we first select all state pairs $(s_{i}, s_{j}) \in \mathcal{N}_{d_p}(s_c) \times \mathcal{N}_{d_p}(s_g)$. We then take all sequences of transitions $\tau_{ij} = \{s_i, a_i, s_{i+1}, ... , s_{j-1}, a_{j-i}, s_{j}\}$ that start from $s_i$, end at $s_j$, and are below a length threshold in terms of timesteps. We sort them based on $\mathcal{R}(\tau_{ij})$, and return the trajectory with the highest reward. We call this trajectory retrieval process `Perceptual Experience Retrieval' (PER). We use PER only to retrieve short trajectory segments between close-by states $(s_c, s_g)$ (i.e., hence the length threshold on $\tau_{ij}$). These are then stitched together into long global trajectories using the planning algorithms defined in the next section. 

\subsection{Long-Horizon Planning Through Stitching Trajectory Segments}
\label{sec:retro_plan}
This section discusses how PER can be employed for long-horizon planning. Classical sampling-based planning algorithms such as RRT \cite{lavalle1998rapidly} or PRM \cite{kavraki1996probabilistic} connect points sampled from obstacle-free space with line segments in order to build a planning graph. We instead reimagine them as memory search mechanisms by altering their subroutines so that whenever an edge is created, a trajectory is retrieved from the replay buffer through PER (eq.\ref{eq:PER}) and stored in that edge. Our new definitions for these subroutines directly mirror the original ones given in \cite{karaman2011sampling}: \\
\underline{\textit{1) Sampling:}} Sampling originally returns a point from obstacle free space. We instead return a state $s_c$ from the replay buffer $\mathcal{M}$ using any distribution (e.g., uniform, or based on visitation-counts). \\
\underline{\textit{2) Lines and Their Cost:}} The equivalent of drawing a line segment in our framework is retrieving a trajectory $\tau_{\mathcal{M}(s_c, s_g)}$, and its length and cost are $len(\tau_{\mathcal{M}(s_c, s_g)})$ and $-\mathcal{R}(\tau_{\mathcal{M}(s_c, s_g)})$ respectively.
\underline{\textit{3) Nearest State and Neighborhood Queries:}} Given a query point $s_i$, these subroutines return the closest point or a neighborhood of points within a distance, among a set of vertices $V = \{s_j\}$. We preserve these definitions, and only replace the metric from euclidean distance to $len(\tau_{\mathcal{M}(s_c, s_g)})$.
    \begin{align*}
    Nearest(V, s_g) &:= \argmin_{s_c \in V}{ \, len(\tau_{\mathcal{M}(s_c, s_g)})} \\
    Near(V, s_g, r) &:= \{s_c \in V \ | \ len(\tau_{\mathcal{M}(s_c, s_g)}) \leq r\}
    \end{align*}
\underline{\textit{4) Collision Tests:}} Collision tests originally prevent the sampling and line drawing subroutines from intersecting obstacles. Since we are planning in retrospect, any such undesirable event can be handled during PER by adjusting the reward function (i.e., if $\tau$ has such an event, this reflects on $\mathcal{R}(\tau)$).

\begin{algorithm}[th]
\caption{R-PRM (Roadmap Construction)}\label{alg:r_prm}
\begin{algorithmic}[1]
\State \textbf{Input:} $f_\phi, \mathcal{M}$
\State $V \gets \{SampleFree_i\}_{i = 1, ..., num\_vertices}; \: E \gets \emptyset$ \Comment{Initialize vertices and edges}
\ForEach{$s_i \in V$}
    \State $U \gets Near(V, s_i, r) \setminus \{s_i\}$ 
    \ForEach{$s_j \in U$} \Comment{Place PER trajectories in edges}
        \State $E \gets E \cup \{(s_i, s_j): \: \tau_{edge} = \tau_{\mathcal{M}(s_i, s_j)}, \ d_{edge} = -\mathcal{R}(\tau_{\mathcal{M}(s_i, s_j)})\}$ 
    \EndFor
\EndFor
\linebreak
\Return $G = (V, E)$
\end{algorithmic}
\end{algorithm}

\begin{algorithm}[th]
\caption{R-PRM (Trajectory Restitching Given the Constructed Roadmap)}\label{alg:r_prm2}
\begin{algorithmic}[1]
\State \textbf{Input:} $s_c, s_g, G = (V, E), \mathcal{R}(\tau), f_\phi, \mathcal{M}$ 
\ForEach{$s_i \in V$} \Comment{Insert $s_c$ and $s_g$ into the PRM graph}
    \If{$len(\tau_{\mathcal{M}(s_c, s_i)}) \leq r$} \Comment{Place PER trajectories in edges}
        \State $E \gets E \cup \{(s_c, s_i): \: \tau_{edge} = \tau_{\mathcal{M}(s_c, s_i)}, \ d_{edge} = -\mathcal{R}(\tau_{\mathcal{M}(s_c, s_i)})\}$
    \EndIf
    \If{$len(\tau_{\mathcal{M}(s_i, s_g)}) \leq r$} 
        \State $E \gets E \cup \{(s_i, s_g): \: \tau_{edge} = \tau_{\mathcal{M}(s_i, s_g)}, \ d_{edge} =-\mathcal{R}(\tau_{\mathcal{M}(s_i, s_g)})\}$
    \EndIf
\EndFor
\linebreak
\State $\tau_{stitched} \gets \emptyset$
\State $\{s_j\} \gets ShortestPath(s_c, s_g, G, \mathcal{R}(\tau))$ \Comment{Trajectory stitching by dynamic programming}
\For{$0 < i < |\{s_j\}|$} \Comment{Concatenate PER trajectories along the shortest path}
\State $\tau_{stitched} \gets \tau_{stitched} \circ \tau_{\mathcal{M}(s_{i-1}, s_i)}$
\EndFor
\linebreak
\Return $\tau_{\mathcal{M}^{\text{*}}(s_c, s_g)} = \tau_{stitched}$
\end{algorithmic}
\end{algorithm}

Using these subroutines directly in-place of their originals, we reimplement experiential equivalents of PRM, RRT, and RRT*, which we call R-PRM, R-RRT, R-RRT*. We denote the resulting planned trajectory as $\tau_{\mathcal{M}^{\text{*}}(s_c, s_g)}$. Algorithms \ref{alg:r_prm}, \ref{alg:r_prm2} describe R-PRM as an example, and the supplementary contains descriptions for R-RRT, R-RRT*. 

We note two things about our proposed planning algorithms. First, they can optimize any general reward function $\mathcal{R}$. As the number of sampled vertices increases, $\mathcal{R}(\tau_{\mathcal{M}^{\text{*}}(s_c, s_g)})$ gets optimized through dynamic programming (i.e., by minimizing the Bellman error between vertices of the roadmap $G$), therefore employing the same mechanism as classical sampling-based planning algorithms \cite{karaman2011sampling}. Second, they operate on an offline dataset of unlabeled transitions which solely consists of high-dimensional on-board sensory data (e.g. images), \textit{without assuming any auxiliary instrumentation in the environment or oracle information that cannot be sensed by the agent on its own}. They therefore aim to relax the assumptions classical sampling-based planning methods make about what constitutes a model (e.g., replacing a geometric environment model with sensory experience) and what constitutes a state (e.g., enabling search and planning directly over images).

\subsection{Refining Memory Contents via Forming and Executing Plans}
\label{sec:memory_opt}

We iteratively form and execute $\tau_{\mathcal{M}^{\text{*}}(s_c, s_g)}$, and whenever execution is successful, we insert the resulting new trajectories back into $\mathcal{M}$. We note that these new trajectories are not exactly the same as $\tau_{\mathcal{M}^{\text{*}}(s_c, s_g)}$, because $\tau_{\mathcal{M}^{\text{*}}(s_c, s_g)}$ contains approximate mismatches between the endpoints of its stitched trajectory segments due to nearest neighbor retrieval. Forming and executing plans this way creates the following perception-action loop: \textbf{i)} $\mathcal{M}$ with refined contents is used to train a more accurate $Q(s_t, a, s_g)$, \textbf{ii)} a more accurate $Q(s_t, a, s_g)$ creates a more accurate distance metric $d_\phi$, \textbf{iii)} a better $d_\phi$ generates better $\tau_{\mathcal{M}^{\text{*}}(s_c, s_g)}$, \textbf{iv)} better $\tau_{\mathcal{M}^{\text{*}}(s_c, s_g)}$ result in higher frequencies of successful execution to further refine $\mathcal{M}$ (see the supplementary for an algorithmic description).
\section{Related work}
\label{related-work}
\underline{\textit{Self-supervised goal reaching:}} Our approach is closely related to goal-reaching methods that combine learning-based distance-regression with graph search, particularly Semi-parametric Topological Memory (SPTM) \cite{savinov2018semi} and Search on the Replay Buffer (SoRB) \cite{eysenbach2019search}, which we compare to in our experiments. 
The key difference of our approach is that when setting the edges of the planning graph, it retrieves transitions that \textit{actually happened} rather than relying on learned distance regression. This brings two main benefits. First is robustness. Local reachability estimates are susceptible to overestimation when evaluated between pairs of states that are far apart or unreachable. This is because such states rarely occur together and are therefore out of distribution for the distance regression model. This creates `hallucinated' shortcuts in the planning graph that corrupt shortest path queries \cite{eysenbach2019search, emmons2020sparse}. To address this, \cite{savinov2018semi} employs temporally consistent localization and adaptive waypoint selection, while \cite{eysenbach2019search} employs distributional Q-learning and an ensemble of Q-functions.
In our approach, eq.\ref{eq:PER} naturally addresses this problem, since it requires an actual short trajectory in the dataset approximately connecting two states before marking them as close.
The second benefit of our approach is that \emph{it can optimize general reward functions}. This is because it decouples the reachability metric $len(\tau)$ (used in nearest neighbor queries and as a threshold to create edges) from the downstream task reward $\mathcal{R}(\tau)$ (used to set edge distances), unlike previous work. 
\\
\underline{\textit{Image-Based Navigation:}} \cite{shah2021recon, shah2022viking, meng2020scaling} present learning-based navigation systems that incrementally build roadmaps through online operation. Our approach has two main differences: \textbf{i)} it builds a roadmap entirely using raw offline data, therefore allowing applications like multi-robot learning without additional loop-closure mechanisms to fuse graphs from multiple agents, \textbf{ii)} our approach can optimize general reward functions, therefore it is not limited to navigation.\\
\underline{\textit{Robot Motion Planning:}} A common approach to motion planning is to first run a sampling-based planning algorithm \cite{lavalle2006planning, karaman2011sampling}, and then refine the result through trajectory optimization \cite{betts1998survey, betts2010practical, kirk2004optimal} to satisfy constraints \cite{underactuated, pietercourse, choset2005principles}. An important bottleneck is that sampling-based planning algorithms require a precomputed map of the environment, and our approach extends such algorithms in a way that relaxes this requirement by replacing a precomputed map with raw exploration experience.\\
\underline{\textit{SLAM and Geometric Maps:}} SLAM based methods \cite{thrun2002probabilistic} can autonomously construct high-fidelity geometric maps \cite{bujanca2019slambench, grisetti2010tutorial}, therefore alleviating the bottleneck of precomputing environment maps. The downside of such approaches is that they can abstract away useful physical and semantic affordances. For example, a purely geometric map cannot plan a path through a traversable field of tall-grass, while our approach can learn such affordances as long as they are represented in past experiences.

\section{Experiments\protect\footnote{All experiments have a corresponding section in the supplementary providing further implementation details.}}
\label{experiments}

\begin{figure}[t]
  \centering
  \includegraphics[width=\textwidth]{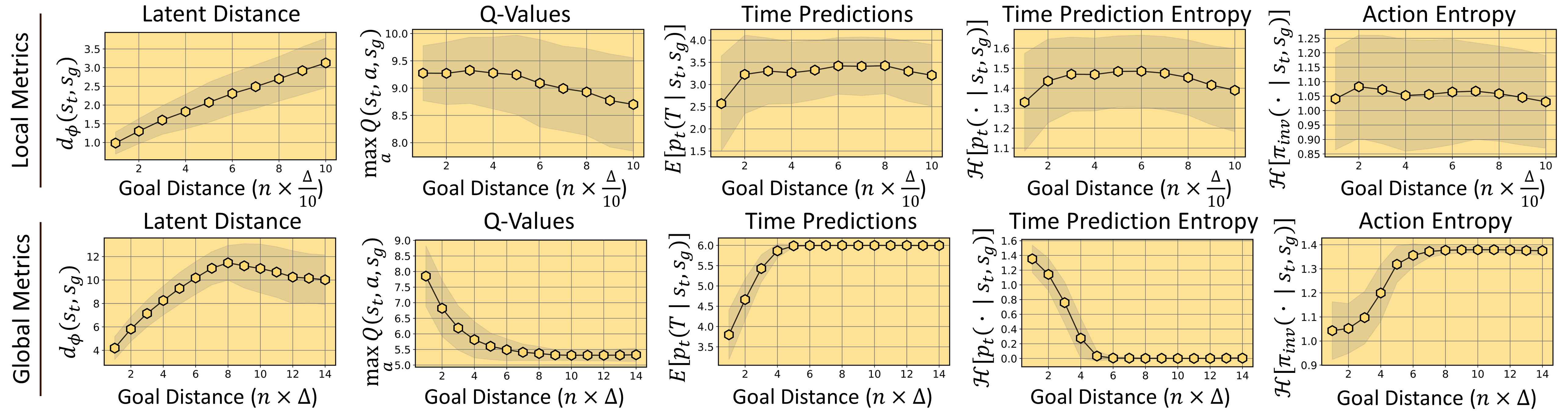}
  \vspace*{-\baselineskip}
  \caption{
  A comparison between perceptual distances $d_{\phi}$ and other suitable metrics from Sec.\ref{sec:architecture}. While all of these metrics are reasonably monotonic with physical reachability (i.e., goal distance), only perceptual distances $d_{\phi}$ do not saturate when evaluated locally (i.e., for close by goals). In addition, the ratio between the variance of $d_{\phi}$ and the slope of its mean is much smaller compared to other sensible metrics (i.e., $d_{\phi}$ has a high signal-to-noise ratio). This means that perceptual distances can implement a more accurate nearest-neighbor criterion for perceptual experience retrieval and trajectory stitching, compared to the other metrics.}
  \label{fig:distances}
  \vspace*{-\baselineskip}
\end{figure}

\underline{\textit{Setup:}} Our experiments are performed in ViZDoom \cite{kempka2016vizdoom}, Habitat \cite{savva2019habitat}, and the Maze2D benchmark \cite{fu2020d4rl}. The VizDoom environment consists of a clover shaped maze. States solely consist of four images $I_{North/East/South/West}$ that form a panorama (i.e., $4 \times 3 \times 160 \times 120$ dimensions), and actions move the agent North/South/East/West by a fixed distance $\Delta$. The maze contains many long-thin column-like obstructions (shown as dots in visualizations). Habitat experiments contain demonstrations on two large-scale scans of real-world apartments: i) Roxboro, with a total area of $~62$ m2, and ii) Annawan, which has a total-area of $~75 m2$. States consist of a single 150 FOV image (i.e., $3 \times 256 \times 256$ dimensions). There are 3 actions: $\{turn\_left\_30\_deg, turn\_right\_30\_deg, move\_forward\_\Delta\}$. Maze2D is a continuous control task, where states consist of the 2D position and velocity of a point mass, and actions correspond to 2D accelerations. In all environments, an offline training dataset is collected by a uniform random walk exploring the environment. For VizDoom and Habitat, this offline training dataset consists of only 300k and 150k timesteps respectively, while for Maze2D there are 1e6 timesteps. Supplementary material contains further details.


\subsection{Experiments in Vizdoom}
\label{sec;vizdoom}

\underline{\textit{Validating Perceptual Representations:}} Fig.\ref{fig:distances} shows that $d_{\phi}(s_t, s_g)$ obtained from our model captures a suitable notion of local reachability. Fig.\ref{fig:local_policy} in turn shows that retrieving nearest neighbor states $\mathcal{N}_{d_p}(s_t)$ from $\mathcal{M}$ using $d_{\phi}$ (i.e., NN retrieval) returns physically close states. 

\begin{figure}[t]
  \centering
  \includegraphics[width=\textwidth]{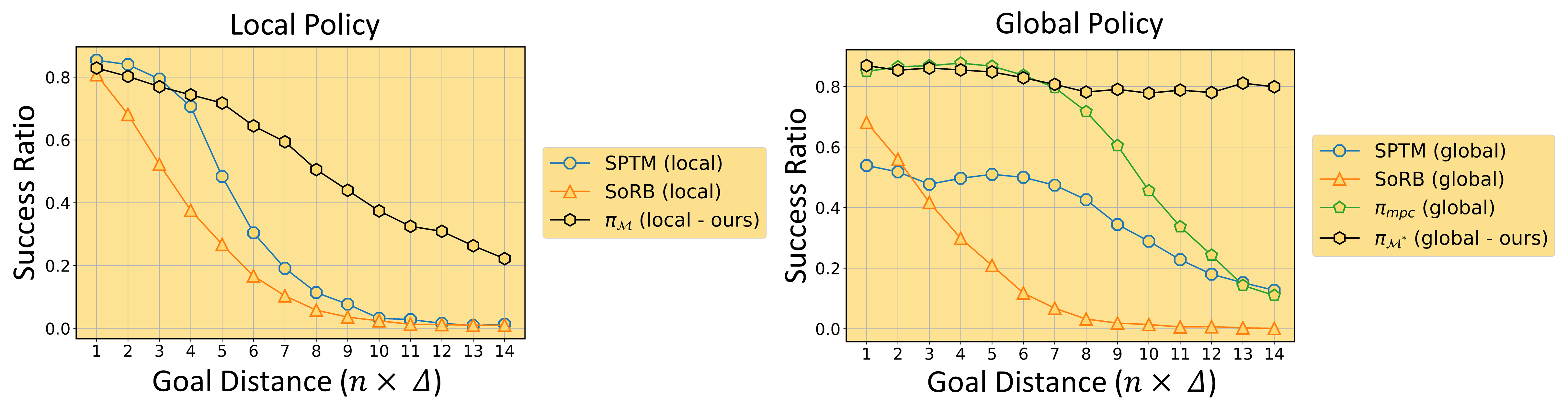}
  \vspace*{-\baselineskip}
  \caption{Comparisons of our local policy $\pi_{\mathcal{M}}$ and global policy $\pi_{\mathcal{M}^{*}}$ with SPTM and SoRB. $\pi_{\mathcal{M}}$ performs well because it avoids getting stuck (as such events are filtered by eq.\ref{eq:PER}), while $\pi_{\mathcal{M}^{*}}$ performs well because it builds robust roadmaps without hallucinated shortcuts; therefore avoiding the main failure modes of the baselines.}
  \label{fig:policy}
\end{figure}

\begin{figure}[t]
  \centering
  \includegraphics[width=\textwidth]{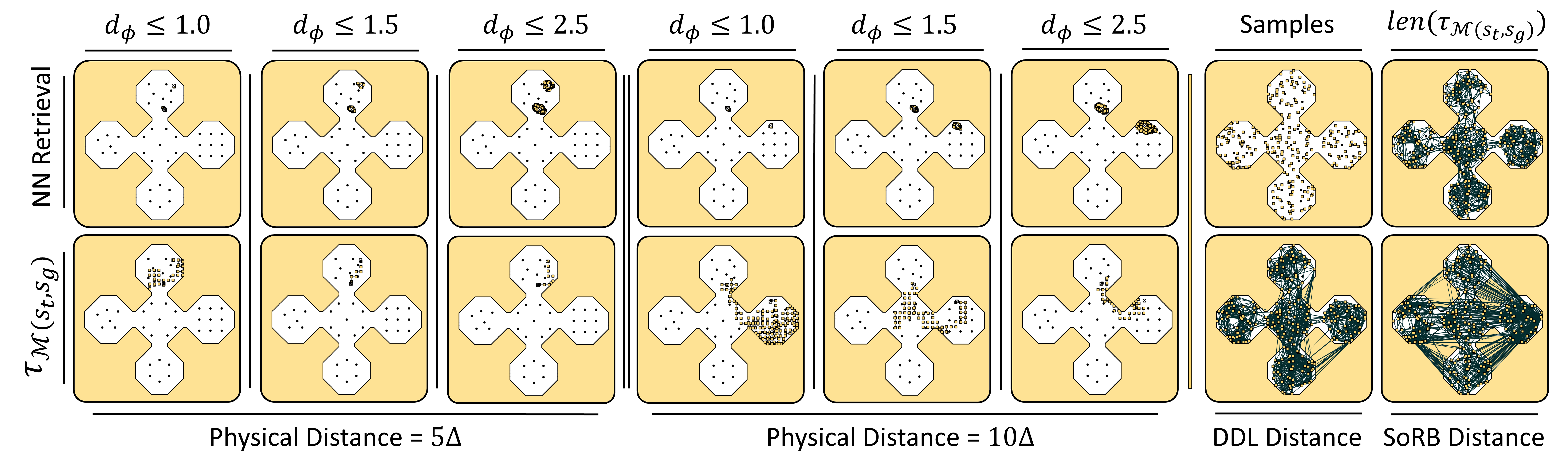}
  \vspace*{-\baselineskip}
  \caption{At the core of PALMER is a process called \textit{perceptual experience retrieval} (PER). Given a query pair of current-goal states, PER searches the replay buffer to retrieve the highest scoring trajectory $\tau_{\mathcal{M}(s_t, s_g)}$ whose first and last states are close to the query pair according to the perceptual distance $d_{\phi}$. \textbf{Left, Middle}: Visualizations of $\tau_{\mathcal{M}(s_t, s_g)}$ retrieved using PER and nearest neighbor states $\mathcal{N}_{d_p}(s_t)$ retrieved using $d_{\phi}$. \textbf{Right}: Setting edges of a roadmap using $len(\tau_{\mathcal{M}(s_t, s_g)})$, compared with distance estimates used in SORB and DDL \cite{hartikainen2019dynamical}. We found that distance estimates from baselines are prone to setting false edges that cross map boundaries.}
  \label{fig:local_policy}
  \vspace*{-\baselineskip}
  \vspace*{-1mm}
\end{figure}

\underline{\textit{Validating Perceptual Experience Retrieval (PER):}} \label{PER_experiments} Fig.\ref{fig:local_policy} shows visualizations of trajectories retrieved with PER. We implement a retrieval policy $\pi_{\mathcal{M}}$ that computes $\tau_{\mathcal{M}(s_t, s_g)}$ through eq.\ref{eq:PER} at each timestep $t$ and executes $argmax_a \ Q(s_t, a, \tau_{\mathcal{M}(s_t, s_g), s, 1})$, therefore forming a model predictive control (MPC) loop. We evaluate $\pi_{\mathcal{M}}$ in an image-based navigation task where start/goal images are sampled randomly to have an euclidean distance $n \times \Delta$ in between,
and a trial is considered successful if the agent can get within $\Delta$ proximity of the goal position within $4 \times n$ time-steps. We use the local policies from SORB \cite{eysenbach2019search} and SPTM \cite{savinov2018semi} as baselines. Fig.\ref{fig:policy} shows the results. The main mode of failure for both SPTM and SORB local policies is that they get stuck in column-like structures. $\pi_{\mathcal{M}}$ avoids this, since eq.\ref{eq:PER} retrieves collision free $\tau_{\mathcal{M}(s_t, s_g)}$.

\underline{\textit{Robust Distances:}} PER also helps avoid hallucinations in local distance regression. Fig.\ref{fig:local_policy} illustrates this point by setting edges between sampled states by thresholding $len(\tau_{\mathcal{M}(s_t, s_g)})$, where methods of \cite{eysenbach2019search, hartikainen2019dynamical} are used as baselines. It can be seen that edges set by $len(\tau_{\mathcal{M}(s_c, s_g)})$ are more robust.

\begin{figure}[t]
  \centering
  \includegraphics[width=\textwidth]{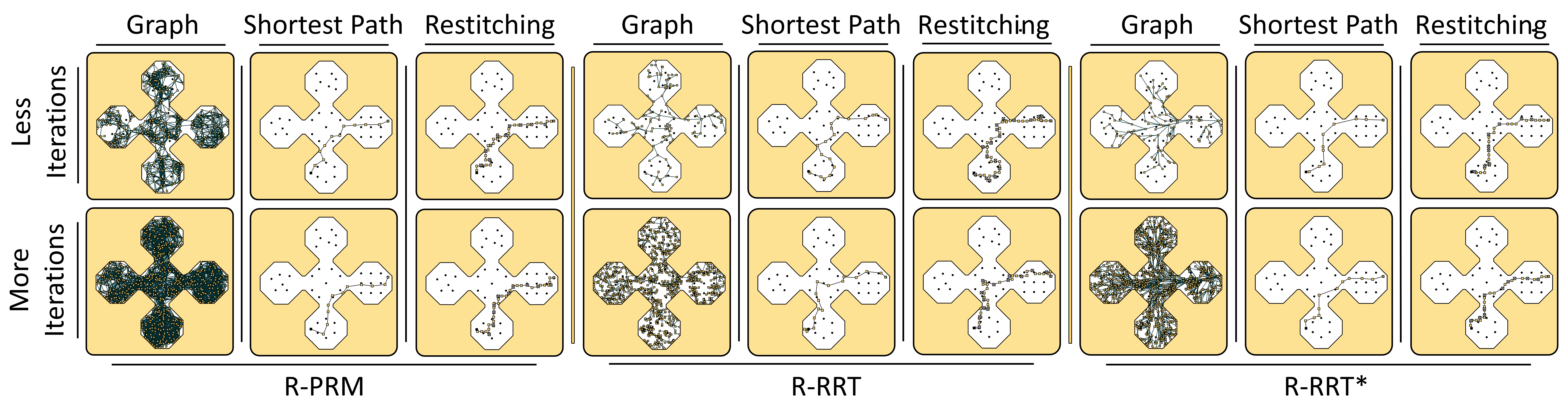}
  \caption{We repurpose conventional sampling-based planning algorithms as memory search mechanisms, by altering their graph building subroutines so that whenever an edge is created a trajectory $\tau_{\mathcal{M}(s_t, s_g)}$ is retrieved through PER and stored in that edge. We visualize the resulting planning graphs produced by our proposed algorithms R-PRM, R-RRT, R-RRT*.}
  \label{fig:global_policy}
\end{figure}

\underline{\textit{Proposed Planning Algorithms:}} Fig.\ref{fig:global_policy} shows visualizations of planning graphs and $\tau_{\mathcal{M}^{\text{*}}(s_c, s_g)}$ produced by R-PRM, R-RRT, and R-RRT*. It can be seen that R-PRM doesn't contain any hallucinated edges, while R-RRT and R-RRT* maintain the visual characteristics of their classical counterparts (i.e., R-RRT has jagged branches with uniform coverage, while R-RRT* has straight branches shooting out from the root). We implement an MPC policy $\pi_{\mathcal{M}^\text{*}}$ that replans at each timestep $t$ using Algorithm \ref{alg:r_prm2} to return $\tau_{\mathcal{M}^{\text{*}}(s_t, s_g)}$, and executes $argmax_a \ Q(s_t, a, \tau_{\mathcal{M}^{\text{*}}(s_t, s_g), s, 1})$.
We again use SORB \cite{eysenbach2019search} and SPTM \cite{savinov2018semi} as baselines.\footnote{For a comparison without confounders, we train SoRB with DDQN \cite{van2016deep} rather than distributional Q-learning \cite{bellemare2017distributional}, and we do not employ temporally consistent localization for SPTM, as such fixes are equally applicable to our method and orthogonal to the discussion. The supplementary provides further elaboration.} Fig.\ref{fig:policy} shows the results. In addition to the local policy getting stuck, a new mode of failure for both baselines is that false distance estimates throw-off graph search by setting hallucinated shortcuts. 
A new baseline is $\pi_{mpc}$, which extends the SPTM local policy by using $p_{fwd}$ and $p_t$ from Sec.\ref{sec:architecture} to implement an MPC loop with n-step look-ahead. $\pi_{mpc}$ avoids getting stuck in columns thanks to n-step lookahead, but still isn't sufficient for global navigation as the accuracy of simulated rollouts from $p_{fwd}$ decreases with the number of timesteps.


\begin{figure}[t]
  \centering
  \includegraphics[width=\textwidth]{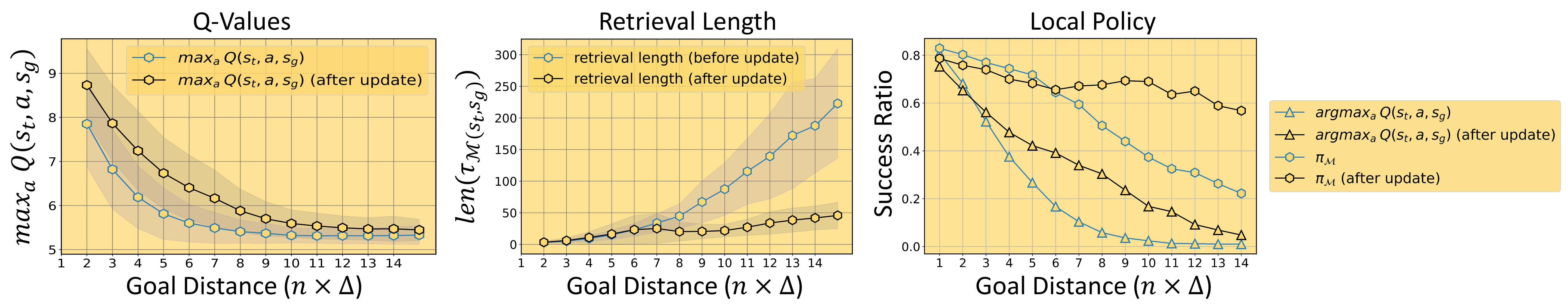}
  \vspace*{-\baselineskip}
  \caption{Memory Refinement: In PALMER, a policy has three groups of parameters: $Q(s_t, a_t, s_g)$, $f_{\phi}$, and the contents of $\mathcal{M}$. Iteratively forming plans through PER and executing them creates a feedback loop between these components, where: i) \textit{actions inform perception} during the training of $f_{\phi}$, ii) \textit{perception facilitates actions} through the formation plans, and iii) \textit{memory serves as the medium} for this reciprocal interaction. As a result, trajectories produced by explicit planning are gradually internalized as implicit behavior encoded in the model parameters. This leads to:  Q-values propagating further into distant goals (\textbf{Left}), memory contents getting closer to optimal (\textbf{Middle}), and performances of local policies showing significant improvement (\textbf{Right}).}
  \label{fig:update}
  \vspace*{-\baselineskip}
\end{figure}
\underline{\textit{Refining Memory Contents:}} We refine the contents of $\mathcal{M}$ by iteratively generating and executing $\tau_{\mathcal{M}^{\text{*}}(s_c, s_g)}$. We then retrain all model components only on the resulting new data that is equal in size to the initial unrefined $\mathcal{M}$.
Fig.\ref{fig:update} shows the results. When $\pi_{\mathcal{M}}$, and $argmax_a Q(a_t, a, s_g)$ are used as policies, their success ratio increases significantly if they are trained on the optimized $\mathcal{M}$. Q-value estimates trained on the optimized $\mathcal{M}$ also propagate better to goals further away. The scaling of $len(\tau_{\mathcal{M}(s_c, s_g)})$ with goal-distance changes from an exponential trend to an approximately linear one, due to the inclusion of transitions from successfully executed $\tau_{\mathcal{M}^{\text{*}}(s_c, s_g)}$. These results highlight that refining memory contents improves the quality of future plans.

\subsection{Experiments in Habitat}
\label{sec:habitat}

\begin{figure}[t]
  \centering
  \includegraphics[width=\textwidth]{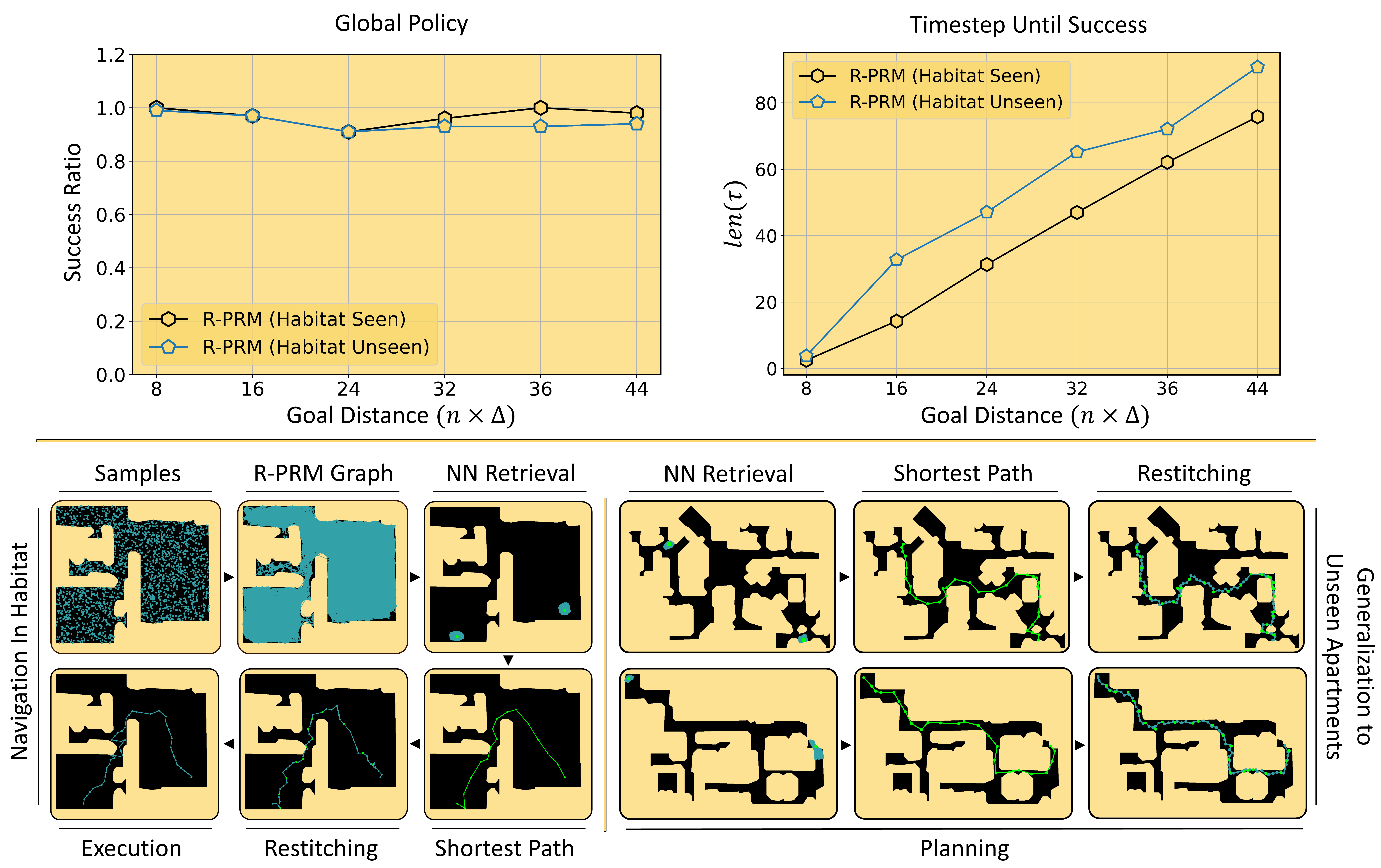}
  \vspace*{-\baselineskip}
  \caption{We evaluate our R-PRM based policy $\pi_{\mathcal{M}^*}$ in the Habitat simulator for image-based navigation. \textbf{Top Left:} Success ratios in training and test apartments. \textbf{Top Right:} Number of timesteps until reaching the goal. ("Habitat seen" refers to the training apartment Roxbox, while "habitat unseen" refers to the test apartment Annawan. \textbf{Bottom:} We found that training the perception model $f_{\phi}$ on a \textit{single} apartment generalizes sufficiently well to allow perceptual experience retrieval and trajectory stitching in any unseen apartment.}
  \label{fig:habitat}
  \vspace*{-\baselineskip}
  \vspace*{-2mm}
\end{figure}


As shown in Fig.\ref{fig:habitat}, we find that our method allows image-based navigation in this new domain with significantly different visuals and layouts (i.e., real-world apartments), action space (i.e., turn-left, turn-right, go-forward), and state space (i.e., single $256 \times 256$  RGB images with 150 FOV). Perhaps more surprisingly, we find that training $f_{\phi}$ only on exploration data from a \textit{single} apartment generalizes substantially well to any unseen apartment, which directly allows perceptual experience retrieval and trajectory stitching when provided with a corresponding replay buffer. For a quantitative evaluation, we randomly pick two apartments, named Roxbox and Annawan. In both apartments, we collect an exploration dataset using a uniform random walk sequence of only 150k timesteps. We train the model components solely on data from Roxbox. We then use them to implement our $\pi_{\mathcal{M}^*}$ policy from Sec.\ref{sec;vizdoom}, which we then evaluate on both apartments. For  $n \in \{8, 16, 24, 32, 36, 44\}$, we randomly sample $100$ pairs of start and goal-states in a way that the geodesic distance between them lies within $n \times \Delta$ and $(n+8) \times \Delta$ through rejection sampling. A policy is considered successful if it can get within $2 \times \Delta$ proximity of the goal-state. We do not plot the SPTM and SORB baselines, because we found that the models $\pi_{inv}(a | s_t, s_g)$ and $argmax_a \ Q(s_t, a, s_g)$ that they use as local navigation policies achieved almost zero percent success rate in reaching local goals beyond $\sim 2 \times \Delta$ distance. We empirically observed that most of the time these policies get stuck in repetitive rotational motions without moving forward. This is most likely due to the difficulty of offline RL training with hindsight relabelling over random-walk data obtained with a much more challenging non-cartesian action space $\{turn\_left\_30\_deg, turn\_right\_30\_deg, move\_forward\_\Delta\}$.

\subsection{Experiments in Maze2D}
\begin{table}[H]
\resizebox{\columnwidth}{!}{%
\begin{tabular}{|c|c|c|c|c|c|c|c|c|c|}
\hline
               & SAC   & SAC-off & BEAR   & AWR  & BCQ    & CQL  & IQL  & Diffuser & PALMER \\ \hline
maze2d-umaze   & 110.4 & 145.6   & 28.6   & 25.2 & 41.5   & 31.7 & 89.6 & \textbf{182.1}    & 131.76 \\ \hline
maze2d-medium  & 69.5  & 82.0    & 89.8   & 33.2 & 35.0   & 26.4 & 105.2 & 332.9    & \textbf{416.28} \\ \hline
maze2d-large   & 14.1  & 1.5     & 19.0   & 70.1 & 23.2   & 40   & 159.9 & 328.1    & \textbf{361}    \\ \hline
\end{tabular}
}
\caption{Total rewards on the Maze2D benchmark, which is a continuous control task that requires long-horizon planning. Our R-PRM based $\pi_{\mathcal{M}^*}$ policy achieves comparatively strong performance.}
\vspace{-\baselineskip}
\vspace{-\baselineskip}
\label{tab:maze2d}
\end{table}

To test our method on a continuous control task, we perform additional experiments on the Maze2D benchmark. As shown in Table.\ref{tab:maze2d}, we find that the same $\pi_{\mathcal{M}^*}$ policy from sections \ref{sec;vizdoom} and \ref{sec:habitat} achieves strong performance, and can solve mazes of all three complexities.

\section{Discussion and Future Directions} \label{discussion}
\underline{\textit{Is PALMER less expressive than standard deep Q-learning:}} Two important premises of deep Q-learning \cite{sergeycourse, pietercourse} are: i) minimizing Bellman error through temporal-difference (TD) updates can restitch observed transitions in new optimal ways \cite{fu2020d4rl, chebotar2021actionable}, ii) a neural network can learn to extrapolate Q-values to unobserved but close-by states in high-dimensional spaces (e.g. images) \cite{mnih2015human}. Both arguments are equally valid for our approach, since it can: i) restitch transitions at arbitrary resolutions (i.e., anywhere from one-step transitions to multi-step trajectories) by virtue of sampling-based planning, ii) group together close-by states through $d_{\phi}$. Therefore, PALMER is an RL algorithm that: \textbf{i)} optimizes Bellman error through sampling-based optimal planning rather than gradient-based TD-updates \cite{mnih2015human}, \textbf{ii)} performs extrapolation between states using a perceptual-backbone $f_{\phi}$ rather than a deep Q-network, and \textbf{iii)} replaces the greedy-policy $argmax_a \ Q(s_t, a, s_g)$ and value estimate $max_a \ Q(s_t, a, s_g)$ with $argmax_a \ Q(s_t, a, \tau_{\mathcal{M}^{\text{*}}(s_t, s_g), s, 1})$ and $\mathcal{R}(\tau_{\mathcal{M}^{\text{*}}(s_c, s_g)})$ respectively. The key benefits of these alterations come into play when $s_t$ and $s_g$ are far apart, and these benefits are: \textbf{i)} the PER mechanism in eq.\ref{eq:PER} that prevents hallucinations in $Q(s_t, a, s_g)$, \textbf{ii)} global propagation of value estimates by virtue of employing sampling-based planning methods, which are known to be particularly proficient at searching high-dimensional state spaces across long-horizons \cite{choset2005principles, lavalle2006planning}.\\
\underline{\textit{Combining PALMER with standard deep Q-learning:}} Our approach can also be flexibly combined with any traditional Q-learning method \cite{mnih2015human, hessel2018rainbow, haarnoja2018soft}, by using our proposed planning algorithms (Sec.\ref{sec:retro_plan}) as experience replay methods \cite{schaul2015prioritized}. This alternative approach stitches together $\tau_{\mathcal{M}^{\text{*}}(s_c, s_g)}$ during training, and perform backwards TD-updates over this trajectory starting from $s_g = \tau_{\mathcal{M}^{\text{*}}(s_c, s_g), s, -1}$ and ending at $s_c = \tau_{\mathcal{M}^{\text{*}}(s_c, s_g), s, 0}$. As suggested by Fig.\ref{fig:update}, this can allow value estimates $Q(s_t, a, s_g)$ to propagate more globally. Our proof-of-concept experiments identify this as a promising direction, and we leave a further extensive evaluation to future work.
\underline{\textit{Connections to contingency learning:}} Contingency learning refers to the acquisition of knowledge of statistical correlations between percepts \cite{skinner2014contingencies, o2001sensorimotor, noe2004action}. Following this definition, \textit{PALMER can be interpreted as a contingency learning framework, as the latent distance metric $d_\phi$ captures statistically how likely two states are to be observed in close temporal proximity}. The knowledge of these statistical contingencies between states is then used for long-horizon decision making through the proposed perceptual experience retrieval and planning mechanisms.

\section{Conclusion and Limitations}
\label{discussion}

We presented PALMER, a long-horizon planning method that combines learning-based perceptual representations with classical sampling-based planning algorithms. Given a goal state $s_g$ and reward function $\mathcal{R}$, our method searches the contents of an offline replay-buffer $\mathcal{M}$ to stitch together a sequence of transitions $\tau_{\mathcal{M}^{\text{*}}(s_c, s_g)} = \{s_1, a_1, s_2, ...\}$ that reaches $s_g$ while maximizing $\mathcal{R}$. This results in an experiential framework for long-horizon planning that is significantly more robust and sample efficient compared to baselines.


Our experiments show that PALMER can successfully solve long-horizon planning tasks from continuous high-dimensional inputs. In particular, we have shown that given an offline dataset of only 150k transitions (i.e., compared to sample complexities around the orders of magnitude 1e6-1e7 common in RL) obtained from an entirely uniform random-walk (i.e., which is significantly less structured compared to on-policy rollouts), it allows image-based navigation between any two points in large-scale scans of real-world apartments.

We believe that our memory-based planning perspective highlights a number of interesting questions for future research. First, which transitions should be kept in the replay buffer $\mathcal{M}$, and which ones should be discarded? $\mathcal{M}$ cannot be infinitely expanded after deployment, and it is critical to distill away redundancies between stored experiences. Second, when the environment undergoes a change, which transitions in the replay buffer remain valid and can still be used for planning, and which ones become invalid? A mechanism that can answer this question can allow quick and sample-efficient adaptation to environmental changes. Third, how can we extend $f_{\phi}$ to allow more abstract associations and functional equivariances between states? This can improve generalization by defining a more flexible notion of experience retrieval that can recycle past behavior in new contexts and for new tasks. We leave these questions to future work.


\bibliographystyle{IEEEtran}
\bibliography{palmer_arxiv}

\newpage
\section*{Checklist}


\begin{enumerate}

\item For all authors...
\begin{enumerate}
  \item Do the main claims made in the abstract and introduction accurately reflect the paper's contributions and scope?
    \answerYes{}
  \item Did you describe the limitations of your work?
    \answerYes{}
  \item Did you discuss any potential negative societal impacts of your work?
    \answerNA{}
  \item Have you read the ethics review guidelines and ensured that your paper conforms to them?
    \answerYes{}
\end{enumerate}

\item If you are including theoretical results...
\begin{enumerate}
  \item Did you state the full set of assumptions of all theoretical results?
    \answerNA{}
        \item Did you include complete proofs of all theoretical results?
    \answerNA{}
\end{enumerate}

\item If you ran experiments...
\begin{enumerate}
  \item Did you include the code, data, and instructions needed to reproduce the main experimental results (either in the supplemental material or as a URL)?
    \answerYes{}
  \item Did you specify all the training details (e.g., data splits, hyperparameters, how they were chosen)?
    \answerYes{}
        \item Did you report error bars (e.g., with respect to the random seed after running experiments multiple times)?
    \answerYes{}
        \item Did you include the total amount of compute and the type of resources used (e.g., type of GPUs, internal cluster, or cloud provider)?
    \answerYes{}
\end{enumerate}

\item If you are using existing assets (e.g., code, data, models) or curating/releasing new assets...
\begin{enumerate}
  \item If your work uses existing assets, did you cite the creators?
    \answerYes{}
  \item Did you mention the license of the assets?
    \answerNA{}
  \item Did you include any new assets either in the supplemental material or as a URL?
    \answerNo{}
  \item Did you discuss whether and how consent was obtained from people whose data you're using/curating?
    \answerNA{}
  \item Did you discuss whether the data you are using/curating contains personally identifiable information or offensive content?
    \answerNA{}
\end{enumerate}

\item If you used crowdsourcing or conducted research with human subjects...
\begin{enumerate}
  \item Did you include the full text of instructions given to participants and screenshots, if applicable?
    \answerNA{}
  \item Did you describe any potential participant risks, with links to Institutional Review Board (IRB) approvals, if applicable?
    \answerNA{}
  \item Did you include the estimated hourly wage paid to participants and the total amount spent on participant compensation?
    \answerNA{}
\end{enumerate}

\end{enumerate}

\newpage

\appendix
\section*{Supplementary Material}
This document presents additional implementation details, visualizations, and conceptual discussions that were excluded or briefed due to space limitations in the main paper. The organization mirrors the sections from the main paper (with the exact same order and numbering), and the exposition generally maintains a question-answer format. It aims to provide significantly more details, at a degree of clarity sufficient for re-implementation. As such, we  recommend referring to this document whenever any section in the main paper can benefit from further elaboration. The supplementary material consists of: 
\begin{itemize}
    \item {\small \fontfamily{qcr}\selectfont vizdoom\_nav} --- navigation videos in VizDoom.
    \item {\small \fontfamily{qcr}\selectfont habitat\_roxbox\_nav} --- navigation videos in the training apartment Roxbox in Habitat.
    \item {\small \fontfamily{qcr}\selectfont habitat\_annawan\_nav} --- navigation videos in the test apartment Annawan in Habitat.
    \item {\small \fontfamily{qcr}\selectfont maze2d\_experiments} --- goal reaching videos in Maze2D.
    \item {\small \fontfamily{qcr}\selectfont planning\_vis} ---  videos visualizing the execution of R-RRT and R-RRT*.
    \item {\small \fontfamily{qcr}\selectfont code} --- source code for PALMER.
\end{itemize}

\subsection*{\underline{Table of Contents}}

\textbf{1. Introduction}
\begin{itemize}
\item \hyperlink{q1}{Why learn a latent distance metric for nearest neighbor retrieval?}
\item \hyperlink{q2}{What does stitching transitions together mean?}
\end{itemize}
\textbf{2. Perception-Action Loop with Memory Reorganization}

\textbf{\quad 2.1. Perceptual Representations that Capture Local Reachability}
\begin{itemize}
\item \hyperlink{q3}{What does local reachability mean?}
\end{itemize}
\textbf{\quad 2.2. Representation Learning via Reinforcement Learning}
\begin{itemize}
\item \hyperlink{q4}{How are the model components trained, and what are their exact inputs and outputs?}
\item \hyperlink{q5}{What do the terms in the contrastive loss-function $L_Q$ mean?}
\end{itemize}
\textbf{\quad 2.3. Perceptual Experience Retrieval (PER)}
\begin{itemize}
\item \hyperlink{q6}{How do we solve the optimization problem in equation 1 from the main paper?}
\item \hyperlink{q7}{What do the retrieved trajectories $\tau_{\mathcal{M}(s_c, s_g)}$ look like?}
\end{itemize}
\textbf{\quad 2.4. Long-Horizon Planning Through Stitching Trajectory Segments}
\begin{itemize}
\item \hyperlink{q8}{How does R-PRM work in detail?}
\item \hyperlink{q9}{How does R-RRT work in detail?}
\item \hyperlink{q10}{How does R-RRT* work in detail?}
\item \hyperlink{q11}{What does optimizing the Bellman error on the roadmap mean?}
\item \hyperlink{q12}{What does restitching transitions at arbitrary resolution mean?}
\end{itemize}
\textbf{\quad 2.5. Refining Memory Contents via Forming and Executing Plans}
\begin{itemize}
\item \hyperlink{q13}{What does optimizing memory contents mean?}
\item \hyperlink{q14}{How does the perception-action loop in PALMER work in detail?}
\end{itemize}
\textbf{3. Related Work}
\begin{itemize}
\item \hyperlink{q15}{What is the main reason why memory-based reasoning over actually observed transitions is necessary? Why are methods like SPTM or SoRB that solely rely on learning-based distance estimates
are inherently prone to false predictions?}
\item \hyperlink{q16}{What are the details for our implementations of SoRB and SPTM?}
\end{itemize}
\textbf{4. Experiments}\\

\textbf{\quad Setup}
\begin{itemize}
    \item \hyperlink{q17}{What are the details for the experimental setup in VizDoom?}
\end{itemize}
\textbf{\quad Validating Perceptual Experience Retrieval (PER)}
\begin{itemize}
    \item \hyperlink{q19}{What is the exact evaluation process that produced Fig.4 in the main paper?}
\end{itemize}
\textbf{\quad Robust Distances}
\begin{itemize}
    \item \hyperlink{q20}{What is the exact evaluation process for the right panel of Fig.5 in the main paper?}
\end{itemize}
\textbf{\quad Proposed Planning Algorithms}
\begin{itemize}
    \item \hyperlink{q21}{Why does the policy $\pi_{\mathcal{M}^*}$ use R-PRM for planning?}
    \item \hyperlink{q22}{What are the details for the $\pi_{mpc}$ baseline?}
\end{itemize}
\textbf{\quad Experiments in Habitat}
\begin{itemize}
    \item \hyperlink{q23}{What are the details for the experimental setup in Habitat?}
    \item \hyperlink{q27}{Why does the agent occasionally take random-looking actions in the habitat navigation trials?}
\end{itemize}
\textbf{5. Discussion and Future Directions}
\begin{itemize}
    \item \hyperlink{q25}{How is PALMER related to the "Options Framework (Sutton et al.)" and "Skill-Chaining (Konidaris et al.)"?}
    \item \hyperlink{q26}{How is PALMER related to "LQR-Trees (Tedrake et al.)?}
\end{itemize}

\subsection*{1 \quad \: Introduction}
\hypertarget{q1}{\underline{\textit{Why learn a latent distance metric for nearest neighbor retrieval:}}} In a low-dimensional state space such as 3D positions, L2 distance (i.e., euclidean distance) directly correlates with \emph{local} physical-reachability (i.e., we emphasize local, because euclidian distances still do not match geodesic distances globally). Therefore in such state-spaces, grouping together two nearby states and treating them as the same single state for downstream global planning should still result in a feasible planned trajectory. By feasible, we mean that the gaps and approximations introduced by state grouping are functionally inconsequential and can be handled reasonably well by a local policy tracking the global planned trajectory. This property doesn't hold in high-dimensional state spaces such as images, since the L2 distance doesn't correlate with physical reachability. The main purpose of $f_{\phi}$ is to project such high-dimensional spaces into a low-dimensional representation space where this property holds, so that nearby states can be fused together to make sampling-based planning computationally tractable.

\begin{figure}[H]
  \centering
  \includegraphics[width=\textwidth]{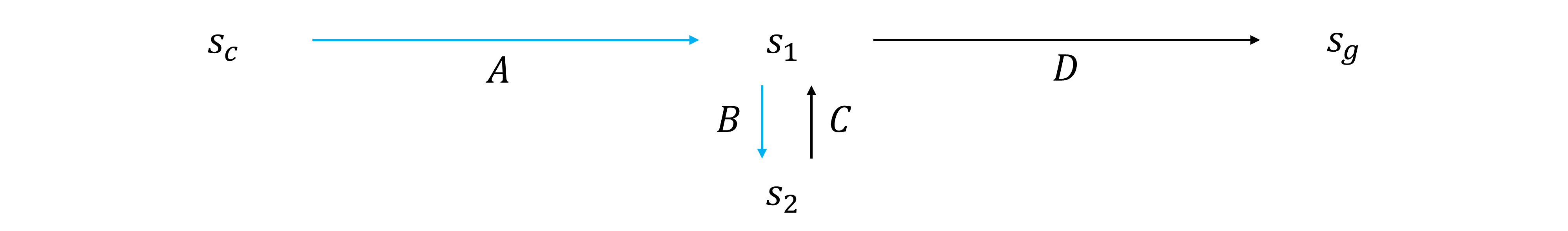}
  \caption{Visualization of stitching together trajectories. If an agent has previous experience of separately going through segments $(A,B)$ and $(C, D)$, it should be able to go from $s_c$ to $s_g$ through the segments $(A, D)$.}
  \label{supfig:2}
\end{figure}

\hypertarget{q2}{\underline{\textit{What does stitching transitions together mean:}}} As shown in Fig\ref{supfig:2}, if there are two separate sequences of transitions in an offline memory buffer that traverse segments $(A,B)$ and $(C, D)$, an agent should be capable of going from $s_c$ to $s_g$ through the segments $(A, D)$ event if such a direct path of transitions was never actually observed. Traditional deep Q-learning methods achieve this by combining and propagating value estimates through TD-updates (i.e., $argmax_a Q(s_1, a, s_g)$ points to segment $D$ after TD updates over the path $(C, D)$, therefore the argmax policy would follow the path $(A, D)$ when going from $s_c$ to $s_g$. \cite{fu2020d4rl} provides a further discussion). In our approach, this is achieved by setting edge distances for $(A, B, C, D)$ in a planning graph through perceptual experience retrieval, and then performing a shortest path computation to retrieve the path $(A, D)$.

\subsection*{2\quad \: Perception-Action Loop with Memory Reorganization}

\subsection*{2.1 \quad \:  Perceptual Representations that Capture Local Reachability}
\hypertarget{q3}{\underline{\textit{What does local reachability mean:}}} At a high-level (and for the special case of image-based navigation) what the term `local reachability' intends to convey is that if two images $I_1$ and $I_2$ are from physically close positions, $d_{\phi} = |f_{\phi}(I_1) - f_{\phi}(I_2)|$ should be small. This in turn provides a metric for grouping together states that is better than the L2 distance in image space, which has no correlation with physical reachability. Such a metric is necessary to make search and sampling-based planning planning over the memory buffer computationally tractable. This learned metric $d_{\phi}$ serves the exact same purpose as the hand-crafted image compression criterion employed in \cite{ecoffet2021first} to initialize cells from states.

\subsection*{2.2 \quad \:  Representation Learning via Reinforcement Learning}

\begin{figure}[H]
  \centering
  \includegraphics[width=\textwidth]{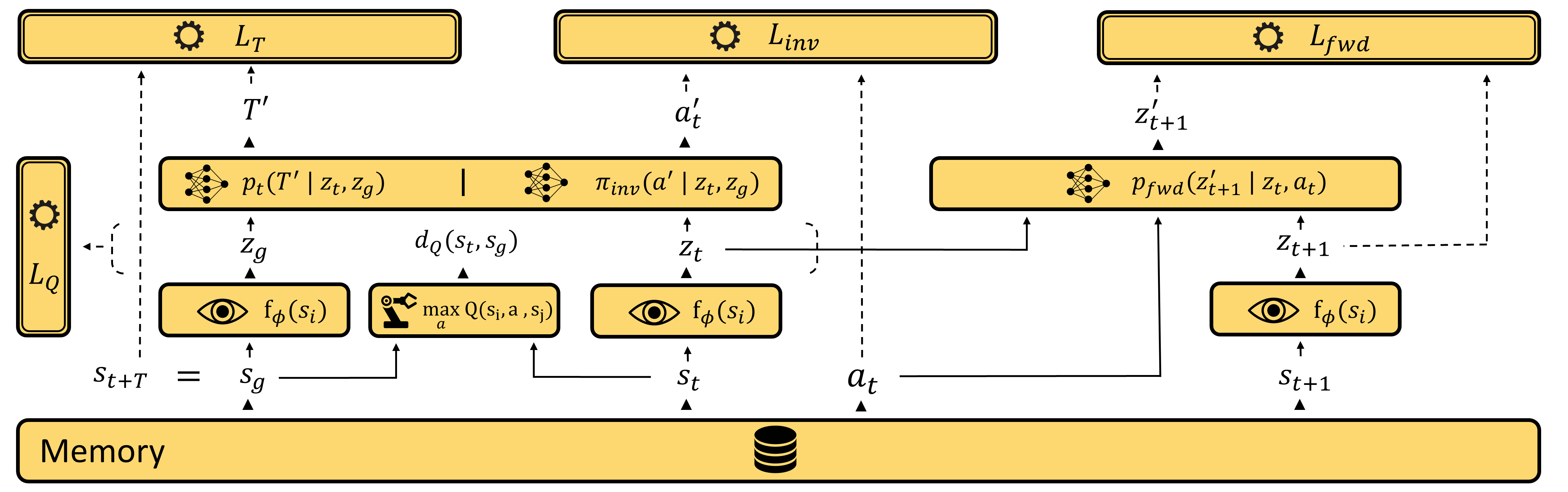}
  \caption{Visualization of the model architecture, reproduced here for ease of reference.}
  \label{supfig:3}
  \vspace*{-2mm}
\end{figure}

\hypertarget{q4}{\underline{\textit{How are the model components trained, and what are their exact inputs and outputs:}}} More detailed descriptions for all model components are given below (Fig.\ref{supfig:3} presents a visualization):

\begin{itemize}[leftmargin=*]
\item The architecture and training of $Q(s_t, a_t, s_g)$ is completely decoupled from the other components. It consists of cascaded convolutional and fully-connected layers with batch normalization and ReLU activations between each layer. It takes as input the concatenated images for current and goal states (i.e., shape $B \times C \times H \times W$), and outputs a vector of Q-values for each action (i.e., shape $B \times num\_actions$). It is trained through offline DDQN \cite{van2016deep} with hindsight goal-relabelling \cite{andrychowicz2017hindsight}. We first sample $t \sim Uniform(0, dataset\_size)$ and $T \sim Geom(p)$, and then retrieve from the replay buffer a transition and a goal state as $(s_t, a_t, s_{t+1}, s_g := s_{t+T})$. We then minimize the TD error $[ \ Q_{\theta}(s_t, a_t,s_g) - (\mathbbm{1}_{s_{t+1}=s_g} + \gamma\mathbbm{1}_{s_{t+1} \neq s_g} \ max_a \ Q_{\theta^{-}}(s_{t+1}, a, s_g)) \ ]^2$, as in \cite{tian2020model}.

\item The perceptual backbone $f_{\phi}(s)$ uses a standard Resnet-18 architecture. It takes as input the images for a given state (i.e., $B \times C \times H \times W$), and outputs a low-dimensional representation vector $z = f_{\phi}(s)$ (i.e., shape $B \times D$). All other components take as input these low-dimensional representations, rather than operating over images.

\item $p_{fwd}(z'_{t+1} \ | \ z_t, a_t)$, $\pi_{inv}(a'_t \ | \ z_t, z_g)$, and $p_{t}(T' \ | \ z_t, z_g)$ all consist of fully-connected layers with ReLU activations. To train them, we first sample $t \sim Uniform(0, dataset\_size)$. We then sample $T$ according to $T \sim Uniform(0, T_{max})$ or $T \sim Uniform(T_{max}, dataset\_size - t)$, half the time from the former distribution, half the time from the latter. We retrieve from the replay buffer a transition and a goal state as $(s_t, a_t, s_{t+1}, s_g := s_{t+T})$, and project them into low dimensional representations $(z_t, a_t, z_{t+1}, z_g)$ using $f_{\phi}$. These are concatenated and passed to models $p_{fwd}(z'_{t+1} \ | \ z_t, a_t), \pi_{inv}(a'_t \ | \ z_t, z_g), p_{t}(T' \ | \ z_t, z_g)$ in a way compatible with their arguments. $p_{fwd}$ outputs the mean for the predicted next state distribution (i.e., shape $B \times D$), and is trained using the MSE loss $L_{fwd}$ with $z_{t+1}$ as the target. $\pi_{inv}$ outputs a vector of probabilities over actions (i.e., shape $B \times num\_actions$), and is trained with the cross-entropy loss $L_{inv}$ with $a_{t}$ as the target. $p_{t}$ also outputs a discrete probability distribution over $[0, T_{max}]$ that predicts the distribution of time-steps to reach the goal, where the last bin $T_{max}$ serves as a catch-all for all values above it. It is trained using the cross entropy loss $L_{T}$, with $T$ as the target. As mentioned in the main paper, all components $f_\phi$, $p_{fwd}(z'_{t+1} \ | \ z_t, a_t)$, $\pi_{inv}(a'_t \ | \ z_t, z_g)$, $p_{t}(T' \ | \ z_t, z_g)$ are trained jointly, with an additional loss function $L_Q$ that regularizes $f_\phi$.

\end{itemize}

\hypertarget{q5}{\underline{\textit{What do the terms in the contrastive loss-function $L_Q$ mean:}}} The loss function $L_Q(s_t, s_g) = l_{hinge}(d_{\phi}(s_t, s_g) - d_{p}) \ \mathbbm{1}_{d_Q(s_t, s_g) \leq c_Q} + l_{hinge}(d_{p} - d_{\phi}(s_t, s_g) ) \ \mathbbm{1}_{d_Q(s_t, s_g) \ge c_Q}$ consists of two penalty terms $l_{hinge}(d_{\phi}(s_t, s_g) - d_{p})$ (i.e., only active when $d_{\phi} \ge d_p$) and $l_{hinge}(d_{p} - d_{\phi}(s_t, s_g) )$ (i.e., only active when $d_{\phi} \leq d_p$). These penalty terms are gated through two complementary indicator functions $\mathbbm{1}_{d_Q(s_t, s_g) \leq c_Q}$ and $\mathbbm{1}_{d_Q(s_t, s_g) \ge c_Q}$. This essentially means that for $L_Q$ to be zero, $d_{\phi} \leq d_p$ should hold (i.e., perceptual representations are close) if and only if $d_Q(s_t, s_g) \leq c_Q$ holds (i.e., states are physically close). The reason for employing such a conservative switching mechanism with a hinge loss in $L_Q$ (i.e., rather than a continuous penalty term as in \cite{chopra2005learning, van2018representation}) is because Q-value estimates are quite inaccurate (especially when $s_c$ and $s_g$ are far apart), and their exact value is generally unreliable (i.e., they can indicate whether two-states are close sufficiently well, but cannot robustly answer how close they are). To pick the hyperparameter $c_Q$, we compute the average Q-value between states in the replay buffer that were observed to be within one-step proximity, and use a fraction of this value to as a conservative estimate. While conceptually the choice for the hyperparameter $d_p$ is arbitrary, we heuristically pick it by examining the average $d_{\phi}$ distance between subsequent states in the replay buffer, obtained from a preliminary $f_{\phi}$ backbone trained without $L_Q$.

\subsection*{2.3 \quad \:  Perceptual Experience Retrieval (PER)}
\hypertarget{q6}{\underline{\textit{How do we solve the optimization problem in equation 1 from the main paper:}}} Our experiments use $-\mathcal{R(\tau)} = len(\tau)$. We first compute the perceptual representations $z$ for all states in the replay buffer, and stack them into a tensor block of shape $(dataset\_size, D)$. Given $s_c$ and $s_g$, we search this tensor with vectorized masking operations to retrieve a set of neighboring states $N(s_c, d_p)$ and $N(s_g, d_p) $ (i.e., sets of states within a perceptual distance threshold $d_p$ of $s_c$ and $s_g$), to address cons.4. We sort the resulting pairs of states $(s_i, s_j) \in N(s_c, d_p) \times N(s_g, d_p)$ in terms of $j - i$, and filter these pairs using cons.5. We then pick the first (i.e., closest) pair, and return all the states with indices between $i, j$ from the replay buffer as the resulting trajectory $\tau_{\mathcal{M}(s_c, s_g)}$. The important thing to emphasize about all of these operations is that they can be trivially vectorized, and therefore the optimization problem in eq.3-5 can be solved in less time than a forward pass of $f_{\phi}$.

\begin{figure}[h]
  \centering
  \includegraphics[width=\textwidth]{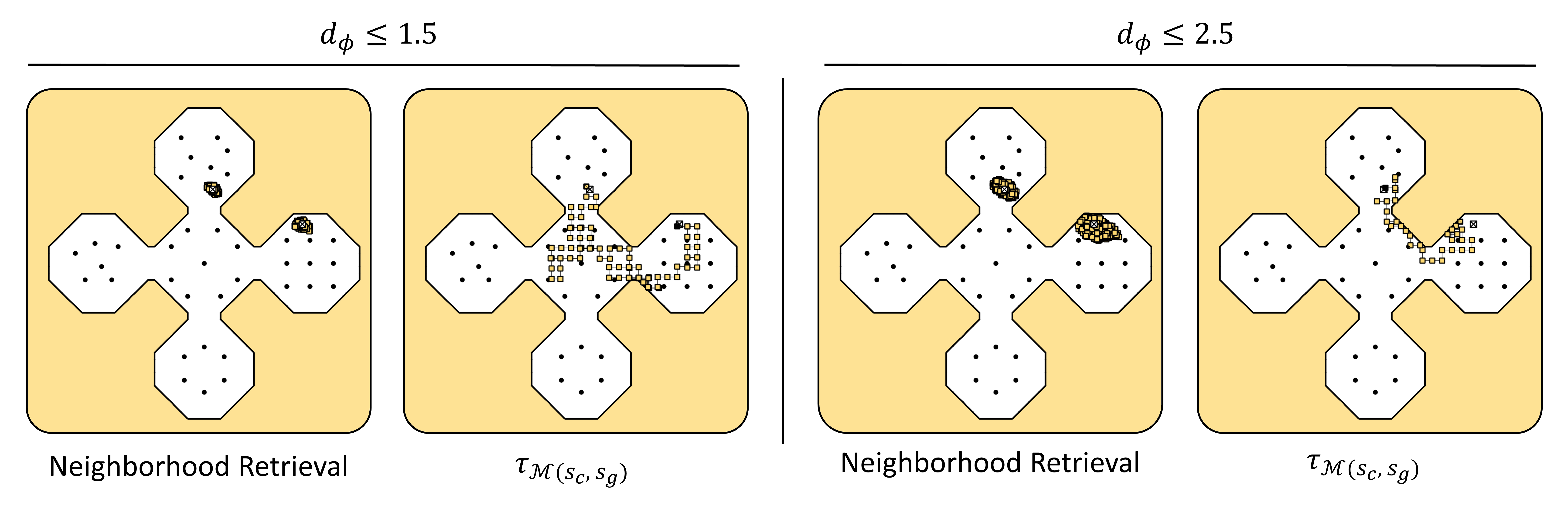}
  \caption{Visualization of retrieved trajectories, with different perceptual distance threshold. Query states $s_c$ and $s_g$ are represented as white squares with a cross in the center, while the start and end points of the retrieved trajectory $\tau_{\mathcal{M}(s_c, s_g)}$ are denoted with a black square and a yellow square with a diagonal dash respectively.}
  \label{supfig:4}
  \vspace*{-2mm}
\end{figure}

\hypertarget{q7}{\underline{\textit{What do the retrieved trajectories $\tau_{\mathcal{M}(s_c, s_g)}$ look like:}}} As the perceptual distance threshold for $d_{\phi}$ increases, the physical radii spanned by the nearest neighbor sets $N(s_c, d_p)$ and $N(s_g, d_p)$ increase, as shown in Fig.\ref{supfig:4}. As a result, constraint 4 in the PER equation gets looser, more trajectories satisfy constraints 4 and 5 (because their start and end points are allowed to deviate further from the query pair $s_c, s_g$), and therefore the optimization in equation 3 returns a shorter trajectory $\tau_{\mathcal{M}(s_c, s_g)}$.

\subsection*{2.4 \quad \:  Long-Horizon Planning Through Stitching Trajectory Segments}

\begin{algorithm}[h]
\caption{Classic PRM (Roadmap Construction)}\label{supalg:1}
\begin{algorithmic}[1]
\State $V \gets \{SampleFree_i\}_{i = 1, ..., num\_vertices}; \: E \gets \emptyset$ \Comment{Initialize vertices and edges}
\ForEach{$s_i \in V$}
    \State $U \gets Near(V, s_i, r) \setminus \{s_i\}$ 
    \ForEach{$s_j \in U$} \Comment{Draw lines as edges}
        \State \algorithmicif \ {$CollisionFree(s_i, s_j)$} \algorithmicthen \ $E \gets E \cup \{(s_i, s_j), (s_j, s_i)\}$ 
    \EndFor
\EndFor
\linebreak
\Return $G = (V, E)$
\end{algorithmic}
\end{algorithm}

\begin{algorithm}[h]
\caption{Classic PRM (Shortest-Path Queries Over the Roadmap)}\label{supalg:2}
\begin{algorithmic}[1]
\State \textbf{Input:} $s_c, s_g, G = (V, E), \mathcal{R}(\tau), f_\phi, \mathcal{M}$ 
\ForEach{$s_i \in V$} \Comment{Insert $s_c$ and $s_g$ into the PRM graph}
    \If{$CollisionFree(s_c, s_i)$} 
        \State $E \gets E \cup \{(s_c, s_i), (s_i, s_c)\}$
    \EndIf
    \If{$CollisionFree(s_i, s_g)$} 
        \State $E \gets E \cup \{(s_i, s_g), (s_g, s_i)\}$
    \EndIf
\EndFor
\linebreak
\Return $\{s_j\} \gets ShortestPath(s_c, s_g, G)$ \Comment{Shortest path through graph search}
\end{algorithmic}
\end{algorithm}

\begin{algorithm}[h]
\caption{R-PRM (Roadmap Construction)}\label{supalg:3}
\begin{algorithmic}[1]
\State \textbf{Input:} $f_\phi, \mathcal{M}$
\State $V \gets \{SampleFree_i\}_{i = 1, ..., num\_vertices}; \: E \gets \emptyset$ \Comment{Initialize vertices and edges}
\ForEach{$s_i \in V$}
    \State $U \gets Near(V, s_i, r) \setminus \{s_i\}$ 
    \ForEach{$s_j \in U$} \Comment{Place PER trajectories in edges}
        \State $E \gets E \cup \{(s_i, s_j): \: \tau_{edge} = \tau_{\mathcal{M}(s_i, s_j)}, \ d_{edge} = -\mathcal{R}(\tau_{\mathcal{M}(s_i, s_j)})\}$ 
    \EndFor
\EndFor
\linebreak
\Return $G = (V, E)$
\end{algorithmic}
\end{algorithm}

\begin{algorithm}[h]
\caption{R-PRM (Trajectory Restitching Given the Constructed Roadmap)}\label{supalg:4}
\begin{algorithmic}[1]
\State \textbf{Input:} $s_c, s_g, G = (V, E), \mathcal{R}(\tau), f_\phi, \mathcal{M}$ 
\ForEach{$s_i \in V$} \Comment{Insert $s_c$ and $s_g$ into the PRM graph}
    \If{$len(\tau_{\mathcal{M}(s_c, s_i)}) \leq r$} \Comment{Place PER trajectories in edges}
        \State $E \gets E \cup \{(s_c, s_i): \: \tau_{edge} = \tau_{\mathcal{M}(s_c, s_i)}, \ d_{edge} = -\mathcal{R}(\tau_{\mathcal{M}(s_c, s_i)})\}$
    \EndIf
    \If{$len(\tau_{\mathcal{M}(s_i, s_g)}) \leq r$} 
        \State $E \gets E \cup \{(s_i, s_g): \: \tau_{edge} = \tau_{\mathcal{M}(s_i, s_g)}, \ d_{edge} =-\mathcal{R}(\tau_{\mathcal{M}(s_i, s_g)})\}$
    \EndIf
\EndFor
\linebreak
\State $\tau_{stitched} \gets \emptyset$
\State $\{s_j\} \gets ShortestPath(s_c, s_g, G, \mathcal{R}(\tau))$ \Comment{Trajectory stitching by dynamic programming}
\For{$0 < i < |\{s_j\}|$} \Comment{Concatenate PER trajectories along the shortest path}
\State $\tau_{stitched} \gets \tau_{stitched} \circ \tau_{\mathcal{M}(s_{i-1}, s_i)}$
\EndFor
\linebreak
\Return $\tau_{\mathcal{M}^{\text{*}}(s_c, s_g)} = \tau_{stitched}$
\end{algorithmic}
\end{algorithm}

\hypertarget{q8}{\underline{\textit{How does R-PRM work in detail:}}} Alg.\ref{supalg:1}, \ref{supalg:2} give step-by-step descriptions for the classical PRM algorithm (adapted from \cite{karaman2011sampling}), while Alg.\ref{supalg:3}, \ref{supalg:4} describe our new definitions for R-PRM. It can be seen that there are two main differences: 
\begin{itemize}[leftmargin=*]
    \item In R-PRM, whenever an edge is created, a trajectory $\tau_{\mathcal{M}(s_c, s_i)}$ is retrieved through perceptual experience retrieval and stored in a field $\tau_{edge}$, while its reward  $-\mathcal{R}(\tau_{\mathcal{M}(s_c, s_i)})$ is stored in a different field $d_{edge}$. 
    
    \item In PRM the length and cost of a line segment are the same (i.e., euclidian distance), whereas in R-PRM the length of a trajectory $len(\tau_{\mathcal{M}(s_c, s_i)})$ and its reward $-\mathcal{R}(\tau_{\mathcal{M}(s_c, s_i)})$ are decoupled. This means that a shortest path query in R-PRM returns a sequence of nodes and edges that optimize the reward function $\mathcal{R}$. An additional step in R-PRM is that at the end of the shortest-path query, all trajectories $\tau_{edge} = \tau_{\mathcal{M}(s_{i-1}, s_i)}$ stored in the returned edges are concatenated into a single trajectory $\tau_{\mathcal{M}^{\text{*}}(s_c, s_g)} = \tau_{stitched}$.
\end{itemize}

\begin{algorithm}[h]
\caption{Classic RRT}\label{supalg:5}
\begin{algorithmic}[1]
\State $V \gets \{s_{init}\}; \: E \gets \emptyset$ \Comment{Initialize vertices and edges}
\For{$i = 1, ..., n$}
    \State $s_{rand} \gets SampleFree_i$ \Comment{Sample random vertex}
    \State $s_{nearest} \gets Nearest(V, s_{rand})$ \Comment{Find the nearest vertex in V}
    \State $s_{new} \gets Steer(s_{nearest}, s_{rand}, r)$ \Comment{Draw a line segment of length $r$}
    \If{$CollisionFree(s_{nearest}, s_{new})$} 
        \State $V \gets V \cup \{s_{new}\} \ ; \ E \gets E \cup \{(s_{nearest}, s_{new}), (s_{new}, s_{nearest})\}$
    \EndIf
\EndFor
\linebreak
\Return $G = (V, E)$
\end{algorithmic}
\end{algorithm}

\begin{algorithm}[h]
\caption{R-RRT}\label{supalg:6}
\begin{algorithmic}[1]
\State $V \gets \{s_{init}\}; \: E \gets \emptyset$ \Comment{Initialize vertices and edges}
\For{$i = 1, ..., n$}
    \State $s_{rand} \gets SampleFree_i$ \Comment{Sample random vertex}
    \State $s_{nearest} \gets Nearest(V, s_{rand})$ \Comment{Find the nearest vertex in V}
    \State $s_{new} \gets \tau_{\mathcal{M}(s_{nearest}, s_{rand}), r}$ \Comment{Get the $r$ 'th state in $\tau_{\mathcal{M}(s_{nearest}, s_{rand})}$ }
    \If{$len(\tau_{\mathcal{M}(s_{nearest}, s_{new})}) \leq r$} 
        \State $V \gets V \cup \{s_{new}\}$
        \State $E \gets E \cup \{(s_{nearest}, s_{new}): \: \tau_{edge} = \tau_{\mathcal{M}(s_{nearest}, s_{new})},$ \\ $ \qquad \qquad \qquad \qquad \qquad \qquad \qquad \quad \: d_{edge} = -\mathcal{R}(\tau_{\mathcal{M}(s_{nearest}, s_{new})})\}$
    \EndIf
\EndFor
\linebreak
\Return $G = (V, E)$
\end{algorithmic}
\end{algorithm}

\hypertarget{q9}{\underline{\textit{How does R-RRT work in detail:}}} Alg.\ref{supalg:5} gives a step-by-step description for the classical RRT algorithm (adapted from \cite{karaman2011sampling}), while Alg.\ref{supalg:6} describes our new definition for R-RRT. In addition to the two previous differences between PRM and R-PRM, there is one additional difference between RRT and R-RRT. In RRT, there is a steering sub-routine that draws a line segment of length $r$ starting from $s_{nearest}$ and extending towards $s_{rand}$, to create a new vertex $s_{new}$. In R-RRT, this is replaced by retrieving a trajectory $\tau_{\mathcal{M}(s_{nearest}, s_{rand})}$ starting from $s_{nearest}$ and ending at $s_{rand}$, and its $r$ 'th state is used to create the new vertex $s_{new}$.

\begin{algorithm}[h]
\caption{Classic RRT*}\label{supalg:7}
\begin{algorithmic}[1]
\State $V \gets \{s_{init}\}; \: E \gets \emptyset$ \Comment{Initialize vertices and edges}
\For{$i = 1, ..., n$}
    \State $s_{rand} \gets SampleFree_i$ \Comment{Sample random vertex}
    \State $s_{nearest} \gets Nearest(V, s_{rand})$ \Comment{Find the nearest vertex in V}
    \State $s_{new} \gets Steer(s_{nearest}, s_{rand}, r)$ \Comment{Draw a line segment of length $r$}
    \linebreak
    \If{$CollisionFree(s_{nearest}, s_{new})$} 
        \State $S_{near} \gets Near(V, s_{new}, r)$
        \State $V \gets V \cup \{s_{new}\}$
        \State $s_{min} \gets s_{nearest}$
        \State $c_{min} \gets Cost(s_{nearest}) + Cost(Line(s_{nearest}, s_{new}))$
        \ForEach{$s_{near} \in X_{near}$} \Comment{Connect along a minimum-cost path}
            \State $c_{near} \gets Cost(s_{near}) + Cost(Line(s_{near}, s_{new}))$
            \If{$CollisionFree(s_{near}, s_{new})$ and $c_{near} \leq c_{min}$}
                \State $s_{min} \gets s_{near} \ ; \ c_{min} \gets c_{near}$
            \EndIf
        \EndFor
        
        \State $E \gets E \cup \{(s_{min}, s_{new})\}$
        \ForEach{$s_{near} \in X_{near}$} \Comment{Rewire the tree}
            \State $c_{near, new} \gets Cost(s_{new}) + Cost(Line(s_{new}, s_{near}))$
            \If{$CollisionFree(s_{new}, s_{near})$ and $c_{near, new} \leq c_{near}$}
                \State $s_{parent} \gets Parent(s_{near})$
                \State $E \gets (E \textbackslash \{ (s_{parent}, s_{near}) \} \cup \{s_{new}, s_{near}\})$
            \EndIf
        \EndFor
    \EndIf
\EndFor
\linebreak
\linebreak
\Return $G = (V, E)$
\end{algorithmic}
\end{algorithm}

\begin{algorithm}[h]
\caption{R-RRT*}\label{supalg:8}
\begin{algorithmic}[1]
\State $V \gets \{s_{init}\}; \: E \gets \emptyset$ \Comment{Initialize vertices and edges}
\For{$i = 1, ..., n$}
    \State $s_{rand} \gets SampleFree_i$ \Comment{Sample random vertex}
    \State $s_{nearest} \gets Nearest(V, s_{rand})$ \Comment{Find the nearest vertex in V}
    \State $s_{new} \gets \tau_{\mathcal{M}(s_{nearest}, s_{rand}), r}$ \Comment{Get the $r$ 'th state in $\tau_{\mathcal{M}(s_{nearest}, s_{rand})}$ }
    \linebreak
    \If{$len(\tau_{\mathcal{M}(s_{nearest}, s_{new})}) \leq r$} 
        \State $S_{near} \gets Near(V, s_{new}, r)$
        \State $V \gets V \cup \{s_{new}\}$
        \State $s_{min} \gets s_{nearest}$
        \State $c_{min} \gets Cost(s_{nearest}) + -\mathcal{R}(\tau_{\mathcal{M}(s_{nearest}, s_{new})})$
        \ForEach{$s_{near} \in X_{near}$} \Comment{Connect along a minimum-cost path}
            \State $c_{near} \gets Cost(s_{near}) + -\mathcal{R}(\tau_{\mathcal{M}(s_{near}, s_{new})})$
            \If{$len(\tau_{\mathcal{M}(s_{near}, s_{new})}) \leq r$ and $c_{near} \leq c_{min}$}
                \State $s_{min} \gets s_{near} \ ; \ c_{min} \gets c_{near}$
            \EndIf
        \EndFor
        
        \State $E \gets E \cup \{(s_{min}, s_{new}): \: \tau_{edge} = \tau_{\mathcal{M}(s_{min}, s_{new})}, d_{edge} = -\mathcal{R}(\tau_{\mathcal{M}(s_{min}, s_{new})})\}$
        \ForEach{$s_{near} \in X_{near}$} \Comment{Rewire the tree}
            \State $c_{near, new} \gets Cost(s_{new}) + -\mathcal{R}(\tau_{\mathcal{M}(s_{new}, s_{near})})$
            \If{$len(\tau_{\mathcal{M}(s_{new}, s_{near})}) \leq r$ and $c_{near, new} \leq c_{near}$}
                \State $s_{parent} \gets Parent(s_{near})$
                \State $E \gets (E \textbackslash \{ (s_{parent}, s_{near}) \} \cup \{s_{new}, s_{near}\})$
            \EndIf
        \EndFor
    \EndIf
\EndFor
\linebreak
\linebreak
\Return $G = (V, E)$
\end{algorithmic}
\end{algorithm}

\hypertarget{q10}{\underline{\textit{How does R-RRT* work in detail:}} Alg.\ref{supalg:7}} gives a step-by-step description for the classical RRT* algorithm (adapted from \cite{karaman2011sampling}), while Alg.\ref{supalg:8} describes our new definition for R-RRT*. R-RRT* almost exactly maintains the tree rewiring machinery employed in RRT*, the only difference being that the line costs (i.e., euclidian distance) are replaced with $-\mathcal{R}(\tau_{\mathcal{M}(s_{i}, s_{j})})$ (i.e., in addition to the previous three differences from R-PRM and R-RRT).

\hypertarget{q11}{\underline{\textit{What does optimizing the Bellman error on the roadmap mean:}}} Dynamic programming based graph-search algorithms like Dijkstra or (discrete) value-iteration generally employ a cost caching mechanism to iteratively update cost-to-come values, and these updates reduce Bellman error (i.e., the difference between the cost-to-come values before and after the update) during forward-search \cite{lavalle2006planning}. R-PRM uses Dijkstra for shortest-path search, while R-RRT* employs the same dynamic programming based tree-rewiring mechanism as the original RRT*, therefore both algorithms are optimizing the Bellman error between the vertices of their graphs through dynamic programming.

\hypertarget{q12}{\underline{\textit{What does restitching transitions at arbitrary resolution mean:}}} The retrospective-planning algorithms R-PRM, R-RRT, R-RRT* are sampling based. This means that if two consequent states $s_t$ and $s_{t+1}$ are retrieved during their $SampleFree_i$ routines, then these algorithms will set an edge using $\tau_{\mathcal{M}(s_{t}, s_{t+1})}$, which is simply the single transition $(s_t, a_t, s_{t+1})$. Therefore, given enough samples, these algorithms can restitch trajectories down to the level of individual transitions.

\subsection*{2.5 \quad \:  Refining Memory Contents via Forming and Executing Plans}

\hypertarget{q13}{\underline{\textit{What does optimizing memory contents mean:}}} A replay buffer $\mathcal{M}$ is a collection of trajectories. Optimizing its contents means adding trajectories to $\mathcal{M}$ that achieve higher total reward.

\begin{algorithm}[h]
\caption{PALMER: Perception-Action Loop with Memory Retrieval}\label{supalg:9}
\begin{algorithmic}[1]
\State \textbf{Input:} $\mathcal{R}(\tau)$
\State $f_\phi.init()$, $Q.init()$, $\mathcal{M} \gets \emptyset$  \Comment{Initialize policy parameters}
\While{$t \leq max\_timestep$}
\newline
\While{$i \ \leq \ num\_exploration\_steps$} \Comment{Exploration}
    \State i) Using [\textit{any suitable method}]: explore the environment
    \Statex \quad \quad \: \: to obtain an exploration trajectory $\tau_{new}$ 
    \State ii) Using [$\tau_{new}$]: \textbf{update} $\mathcal{M}$ \Comment{Memory expansion}
\EndWhile
\newline
\While{$i \ \leq \ num\_updates$}
    \State Using [$\mathcal{M}$]: \textbf{update} $Q(s_t, a_t, s_g)$ \Comment{Train value function}
    \State Using [$\mathcal{M}$ and $Q(s_t, a_t, s_g)$]: \textbf{update} $f_{\phi}(s_t, s_g)$ \Comment{Train perception model}
\EndWhile
\newline
\While{$i \ \leq \ num\_exploitation\_steps$} \Comment{Exploitation}
    \State i) Using [$\mathcal{M}$]: sample a random goal $s_g \sim \mathcal{M}$  
    \State ii) Using [$f_{\phi}(s_c, s_g)$ and $\mathcal{R}(\tau)$]: generate $\tau_{\mathcal{M}^{\text{*}}(s_c, s_g)}$
    \State iii) Using [\textit{any suitable method}]: execute $\tau_{\mathcal{M}^{\text{*}}(s_c, s_g)}$ 
    \Statex \quad \quad \: \: to obtain a real trajectory
    $\tau_{real}$
    \State iv) Using [$\tau_{real}$]: \textbf{update} $\mathcal{M}$ \Comment{Memory optimization}
\EndWhile
\EndWhile
\end{algorithmic}
\end{algorithm}

\hypertarget{q14}{\underline{\textit{How does the perception-action loop in PALMER work in detail:}}} Alg.\ref{supalg:9} gives a step-by-step description of the overall perception-action loop implemented in PALMER. \textbf{What PALMER does is essentially bridging any auxiliary exploration method with any auxiliary exploitation method}, by reorganizing exploration experience in $\mathcal{M}$ into $\tau_{\mathcal{M}^{\text{*}}(s_c, s_g)}$ that can be used for exploitation. The particular way in which the trajectories $\tau_{\mathcal{M}^{\text{*}}(s_c, s_g)}$ can be executed for exploitation has a great deal of flexibility, for example: all actions in $\tau_{\mathcal{M}^{\text{*}}(s_c, s_g)}$ can be executed sequentially in an open-loop manner, the first actions of $\tau_{\mathcal{M}^{\text{*}}(s_c, s_g)}$ generated at each timestep can be executed in a model predictive control (MPC) manner, states in $\tau_{\mathcal{M}^{\text{*}}(s_c, s_g)}$ can be tracked by an auxiliary local feedback controller, or the entirety of $\tau_{\mathcal{M}^{\text{*}}(s_c, s_g)}$ can be used to initialize a separate trajectory optimization method.

\subsection*{3 \quad \:  Related Work}

\hypertarget{q15}{\ul{\textit{What is the main reason why memory-based reasoning over actually-observed transitions is necessary? Why are methods like SPTM or SoRB that solely rely on learning-based distance estimates are inherently prone to false predictions:}}} \label{supsec:critical} By definition, states that are far apart from each other in-terms of physical reachability are rarely observed together in an experiential learning framework (e.g., within the same RL episode, or within close by time-steps during random exploration). Therefore, for any learning-based prediction model that is conditioned on current-goal state pairs (e.g., $Q(s_t, a, s_g), \ \pi_{inv}(a' | s_t, s_g), \ p_t(T' | s_t, s_g)$), if it is trained solely on the observed distribution of experiential data, far apart states will be out of distribution (i.e., they have a low probability of being sampled from the experiential data distribution, therefore they are underrepresented in the replay buffer). This means that predictions for such far apart states will inevitably be inaccurate, unless a hard-negative sampling mechanism is implemented to explicitly push their reachability-estimates lower. The problem with this is that there is no inherent signal solely contained in perceptual input (e.g., images) that can guide the resampling process in a way that is generally applicable to all tasks.

For example, \cite{tian2020model} employs a hard-negative sampling mechanism for a manipulation task using joint pose labels for guidance, but such a mechanism has two bottlenecks: i) it is specific to their particular manipulation task, ii) it assumes auxiliary labels. Similarly, the temporally consistent localization mechanism used in SPTM is essentially a heuristic fix specific to navigation. For the case of SoRB, there is no inherent mechanism in distributional RL that addresses this out-of-distribution issue either. While employing an ensemble of Q-functions could potentially capture the epistemic uncertainty for out-of-distribution pairs to directly address this problem, we used an ensemble of Q-functions in our implementation of SoRB and empirically observed that it was insufficient. A similar conclusion can be drawn from the Fig.8 of the original SoRB paper \cite{eysenbach2019search}, as the bulk of the performance increase appears to be due to distributional RL, and ensembles only provide a moderate benefit.

All of these considerations highlight the importance of memory as a robustification mechanism. To summarize the discussions from above, there are two main reasons that cause false reachability predictions: i) there is no robust and general signal solely contained in the isolated instances of perceptual input $(s_c, s_g)$ (i.e., without the trajectory of states in between that connect them) that can identify whether two states are physically far apart, ii) both physically close and far apart states can occur with a long temporal distance in between. Therefore, an agent needs to rely on memory: in order to identify whether two states are close or not, it should try to remember if it ever actually observed those two states close-by in a segment of past experience.

\hypertarget{q16}{\ul{\textit{What are the details for our implementations of SoRB and SPTM:}}} Tha main difference of our SPTM implementation is that it doesn't employ temporally consistent localization and adaptive waypoint selection. The main differences of our SoRB implementation are: i) we use an ensemble of Q-functions, but they are trained with DDQN rather than distributional RL, ii) we train the Q-function on offlline random-walk data, rather than an online episodic training setup with resets and a reward oracle as employed in the original paper. We acknowledge and emphasize that for SoRB, these differences are the most likely reason for the lower performance level we observed in our evaluations compared to the original paper, as they inevitably reduce the accuracy of Q-values. We however note that our method also employs the same Q-values, and generally these implementation differences in the baselines were chosen to facilitate a clear understanding of our approach without confounders, because: i) they do not directly address the root cause of the false prediction problem (as discussed \hyperlink{q15}{above}), ii) one of the main benefits of our method is that it operates over arbitrary offline data without any resets or reward oracles (and it uses DDQN to train the related Q-function).

\subsection*{4 \quad \:  Experiments}

\subsection*{Setup}

\begin{figure}[h]
  \centering
  \includegraphics[width=\textwidth]{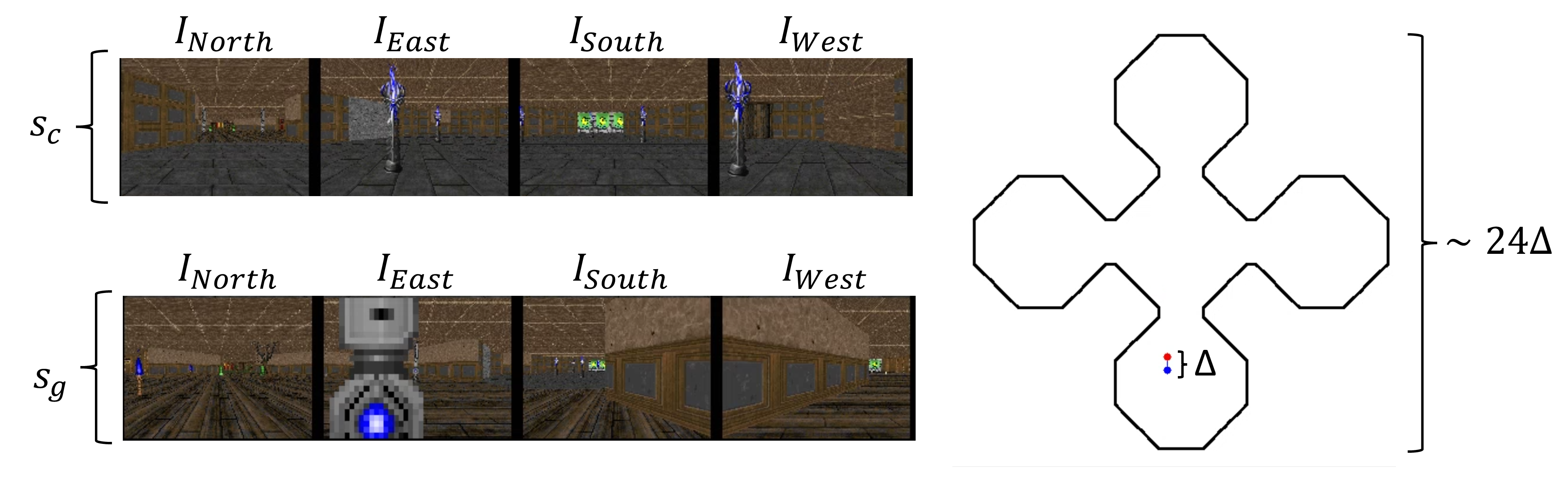}
  \caption{Visualization of the map used in VizDoom experiments. \textbf{Further video visuals of image-based navigation can be found in the folder vizdoom\_nav provided in the supplementary alongside this document}.}
  \label{supfig:5}
  \vspace*{-\baselineskip}
  
\end{figure}

\hypertarget{q17}{\underline{\textit{What are the details for the experimental setup in VizDoom:}}} As shown in Fig.\ref{supfig:5}, states solely consist of four images $I_{North/East/South/West}$ that form a panorama (i.e., $4 \times 3 \times 160 \times 120$ dimensions), and actions move the agent North/South/East/West by a fixed distance $\Delta$. The geodesic distances scale approximately by a factor of $\times 3$ compared to euclidian distances (e.g., If a goal has an euclidian distance of $14 \Delta$, it takes approximately $42$ timesteps for an optimal policy to reach it). The map contains many long-thin column-like obstructions (e.g., torches, pillars, trees), as we found that image-based navigation policies are prone to getting stuck in such obstacles. These obstacles have dynamically changing appearances (e.g., flickering flames on torches, glowing lights on pillars), and can completely block the field of view of the agent after a collision (as shown in $I_{East}$ in Fig.\ref{supfig:5}). The replay buffer $\mathcal{M}$ consists of 300k images obtained from a uniform random walk exploring the map in a single continuous sequence of actions, without resets and rewards.

\subsection*{Validating Perceptual Experience Retrieval (PER)}
\hypertarget{q19}{\underline{\textit{What is the exact evaluation process that produced Fig.4 in the main paper:}}} For every integer value $n \in [0, 14]$, we randomly sample $1000$ pairs of start and goal-states in a way that the euclidian distance between them lies within $n \times \Delta$ and $(n+1) \times \Delta$ through rejection sampling, and a policy is considered successful if it can get within $\Delta$ proximity of the goal-state.

\subsection*{Robust Distances}
\hypertarget{q20}{\underline{\textit{What is the exact evaluation process for the right panel of Fig.5 in the main paper:}}} To produce the roadmap visualizaitons, we randomly sample 250 states from the replay buffer and set the edges between them by thresholding the distance estimates from all methods. Thresholds were calibrated individually and by hand for each baseline, by picking the threshold with the lowest number of false edges until a further reduction in the threshold resulted in splitting the roadmap into a large number of isolated subgraphs (i.e., therefore making it impossible to use it for global planning).

\subsection*{Proposed Planning Algorithms}
\hypertarget{q21}{\underline{\textit{Why does the policy $\pi_{\mathcal{M}^*}$ use R-PRM for planning:}}} PRM and R-PRM are multi-query methods \cite{karaman2011sampling}, meaning that the full roadmap only needs to be constructed once. For any query pair of current-goal states $(s_c, s_g)$, the same roadmap can be reutilized by inserting $(s_c, s_g)$ in the roadmap and performing a shortest path query. In contrast, RRT and RRT* require recreating a full roadmap for every new query pair. Since $\pi_{\mathcal{M}^*}$ replans at each timestep with a different current state $s_t$, PRM based approaches are computationally much cheaper. We also note that planning graphs for all methods in this experiment contain $500$ vertices.

\hypertarget{q22}{\underline{\textit{What are the details for the $\pi_{mpc}$ baseline:}}} The $\pi_{mpc}$ policy uses the $p_{fwd}$ model to obtain simulated rollouts, and uses the $p_t$ model to rank those rollouts in terms of how close they get to the goal. This allows an MPC optimization loop that picks and implements the first action from the most successful simulated rollout. As previously discussed \hyperlink{q15}{above}, the main bottleneck for SPTM and SoRB is the difficulty of estimating reachability metrics solely using the two states $s_c, s_g$ without any consideration of the states in between. Simulated rollouts in $\pi_{mpc}$ naturally address this problem by generating and evaluating entire trajectories. The main bottleneck for $\pi_{mpc}$ is that the accuracy of state predictions in simulated rollouts from $p_{fwd}$ degrade with the rollout length.

\subsection*{Experiments in Habitat}

\begin{figure}[h]
  \centering
  \includegraphics[width=\textwidth]{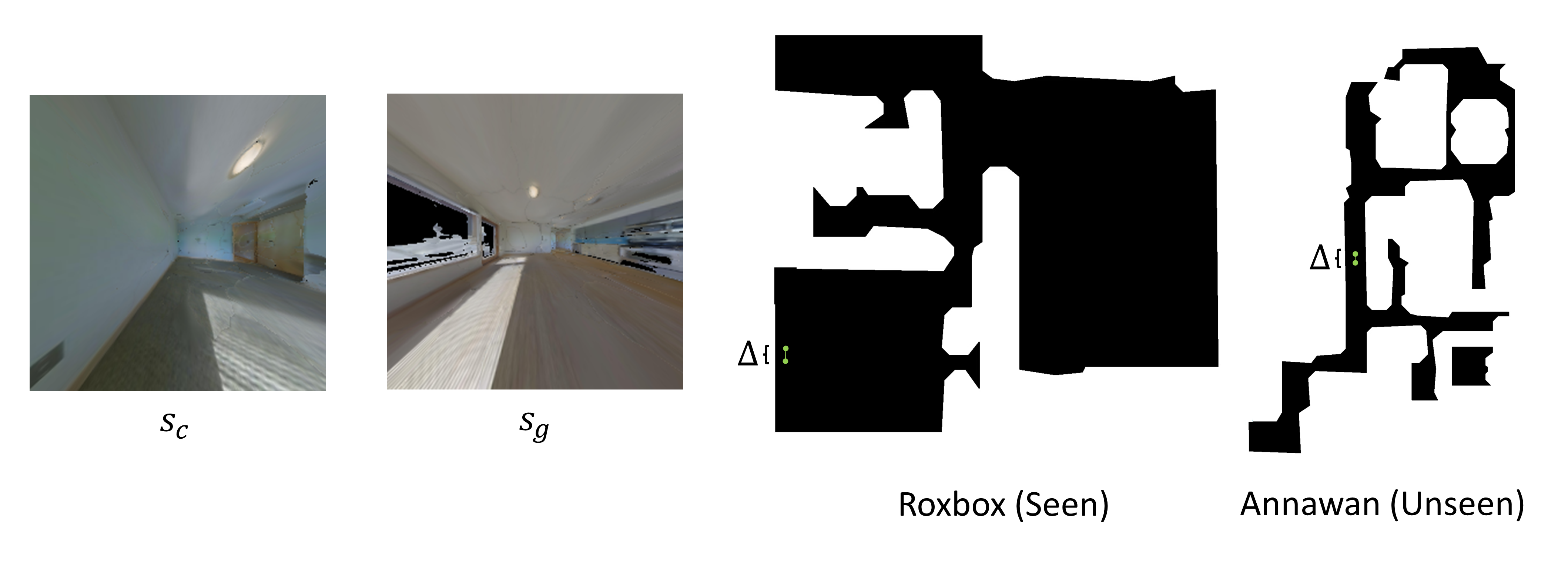}
  \caption{Visualization of the apartments used in Habitat experiments. \textbf{Further video visuals of image-based navigation can be found in the folders habitat\_roxbox\_nav and habitat\_annawan\_nav provided in the supplementary alongside this document}.}
  \label{supfig:6}
  \vspace*{-2mm}
\end{figure}

\hypertarget{q23}{\underline{\textit{What are the details for the experimental setup in Habitat:}}} As shown in Fig.\ref{supfig:6}, states solely consist of a single 150 FOV image (i.e., $3 \times 256 \times 256$ dimensions). There are 3 actions: $\{turn\_left\_30\_deg, turn\_right\_30\_deg, move\_forward\_\Delta\}$. We run evaluations on two randomly picked apartments: Roxbox, and Annawan. In both apartments, we collect a replay buffer $\mathcal{M}$ that consists of 150k images obtained from a uniform random walk exploring the map in a single continuous sequence of actions, without resets and rewards. We use only the memory buffer from Roxbox to train the perception model $f_{\phi}$, and use this same model to do perceptual experience retrieval and trajectory stitching on replay buffers from both apartments. We have observed that the latent distances from $f_{\phi}$ generalize well, and can directly allow perceptual experience retrieval and trajectory stitching without any fine-tuning on the images from the new test apartment.

\hypertarget{q27}{\underline{\textit{Why does the agent occasionally take random-looking actions in the habitat navigation trials:}}} This is due to a combination of two main factors. First, our MPC loop replans from scratch at each timestep using Algorithm.\ref{supalg:4}. This frequent replanning has a destabilizing effect on the control loop, similar to employing a large derivative action in a PID controller (i.e., a strong anticipatory term causes frequent switches in the actions). This first factor is exacerbated by the second main factor: the poor performance of $argmax_a \ Q(s_t, a, s_g)$. This is most likely due to the difficulty of offline RL training with hindsight relabelling over random-walk data of only 150k timesteps obtained with a much more challenging non-cartesian action space $\{turn\_left\_30\_deg, turn\_right\_30\_deg, move\_forward\_\Delta\}$. The restitched trajectories $\tau_{\mathcal{M}^{\text{*}}(s_t, s_g), s, 1})$ produced by R-PRM at each time-step are converted to actions by following their first state $\tau_{\mathcal{M}^{\text{*}}(s_t, s_g), s, 1}$ using $argmax_a \ Q(s_t, a, \tau_{\mathcal{M}^{\text{*}}(s_t, s_g), s, 1})$, and inaccurate Q-values occasionally cause random-looking actions.

There are multiple ways to counter this. The most direct way is to train a better Q-function. Our current Q-function is trained in a particularly challenging setting, as: i) it is trained on entirely offline data with hindsight goal relabelling, ii) this data is collected through uniformly random actions, and iii) it consists of only 150k environment steps. A second way is to instead counter the large derivative action by reducing the replanning rate and introducing a momentum mechanism to the controller. For example, we can replan through R-PRM only every $n$'th timestep (i.e., hence reducing the replanning rate), and act according to $argmax_a \ Q(s_t, a, \tau_{\mathcal{M}^{\text{*}}(s_t, s_g), s, n})$ for the timesteps inbetween (i.e., hence introducing momentum to the controls). We couldn't get this fix to work well, because the Q-values are only accurate up to states that are $\sim 2 \times \Delta$ distance away (hence acting according to $argmax_a \ Q(s_t, a, \tau_{\mathcal{M}^{\text{*}}(s_t, s_g), s, n})$ isn't possible for $n \ge 2$ in our case). We note however that our method can still navigate from any point to any point in a challenging 3D reconstruction of a real-world apartment in Habitat using poor Q-value estimates, highlighting the robustness introduced by memory-based planning.

\subsection*{5 \quad \:  Discussion and Future Directions}

\hypertarget{q25}{\ul{\textit{How is PALMER related to the "Options Framework (Sutton et al.)" and "Skill-Chaining (Konidaris et al.)":}}} The idea of restitching (i.e., chaining) transition sequences from a replay buffer has direct connections to the options framework \cite{sutton1999between} and skill-chaining \cite{konidaris2009skill}. Essentially, PALMER can be thought of as a framework for converting every possible sequence of transitions $\tau \in \mathcal{M}$ in memory into an option $o = \{\pi_o, I_o, \beta_o\}$, where the option policy $\pi_o$ is implemented by simply executing all the actions in $\tau$ in an open-loop manner, the initiation set is $I_o = \{s \in \mathcal{S}: d_\phi(s, \tau_{s, 0}) \leq d_p)\}$, and the termination condition is $\beta_o = \{s \in \mathcal{S}: d_\phi(s, \tau_{s, -1}) \leq d_p)\}$. Therefore, PALMER can be interpreted as a skill-chaining algorithm that converts unstructured transitions in a replay buffer into a set of executable options to be chained.

\hypertarget{q26}{\ul{\textit{How is PALMER related to "LQR-Trees (Tedrake et al.)":}}} A central idea in PALMER is repurposing the edge creation subroutines of sampling-based planning algorithms so that whenever an edge is created some additional processing is done to connect the endpoints (i.e., particularly, perceptual experience retrieval in our case). This approach is directly inspired by the method of LQR-Trees \cite{tedrake2010lqr}, which instead creates a trajectory stabilizing LQR controller to connect the endpoints of each edge. This results in a roadmap of local controllers, rather then a roadmap of memories as in PALMER.



\end{document}